\documentclass{elsarticle}

\usepackage{tikz}
\usepackage{mathtools}
\usepackage{graphicx}
\usepackage{array}
\usepackage{multirow}
\usepackage{siunitx}
\usepackage{amsfonts}
\usepackage{booktabs}
\usepackage{subcaption}
\usepackage{times}
\usepackage{txfonts}
\usepackage{fullpage}

\usetikzlibrary{shapes,arrows,positioning,automata,decorations.text}
\tikzset{
	block/.style = {draw, fill=white, rectangle, minimum height=3em, minimum width=6em},
	input/.style = {coordinate},
	output/.style = {coordinate},
	tmp/.style = {coordinate}
}

\newcommand{\vc}[1]{\mathbf{\boldsymbol{\mathrm{{#1}}}}}

\newcommand{\qud}{\text{Q}}
\newcommand{\trg}{\text{T}}
\newcommand{\cam}{\text{C}}
\newcommand{\grb}{\text{G}}
\newcommand{\ds}{\text{ds}}
\newcommand{\ag}{\text{A}}

\newcommand{\col}{\text{col}}
\newcommand{\lat}{\text{roll}}
\newcommand{\lon}{\text{pitch}}
\newcommand{\yaw}{\text{yaw}}

\newcommand{\Wframe}{\mathcal{W}}
\newcommand{\Aframe}{\mathcal{A}}
\newcommand{\Qframe}{\mathcal{Q}}
\newcommand{\Cframe}{\mathcal{C}}
\newcommand{\Tframe}{\mathcal{T}}

\newcommand{\cphi}{\cos\phi}
\newcommand{\sphi}{\sin\phi}
\newcommand{\ctheta}{\cos\theta}
\newcommand{\stheta}{\sin\theta}
\newcommand{\cpsi}{\cos\psi}
\newcommand{\spsi}{\sin\psi}

\graphicspath{{./Images/}}

\newcommand{\prismkeyword}[1]{\mathtt{#1}}
\newcommand{\prismcomment}[1]{\mbox{\emph{#1}}}
\newcommand{\prismident}[1]{\mathit{#1}}
\newcommand{\prismtab}{\quad}

\definecolor{blue}{rgb}{0,0,1}

\journal{Robotics and Autonomous Systems}

\begin{document}

\begin{frontmatter}

\title{A Continuous-Time Model of an Autonomous Aerial Vehicle to Inform and Validate Formal Verification Methods}

\author[glasgow]{Murray~L.~Ireland\corref{corauthor}}
\cortext[corauthor]{Corresponding author}
\ead{Murray.Ireland@glasgow.ac.uk}
\author[glasgow]{Ruth~Hoffmann}
\author[glasgow]{Alice~Miller}
\author[glasgow]{Gethin~Norman}
\author[sheffield]{Sandor~M.~Veres}

\address[glasgow]{School of Computing Science, University of Glasgow, Glasgow, UK}
\address[sheffield]{Autonomous Systems and Robotics Research Group, University of Sheffield, Sheffield, UK}

\begin{abstract}
If autonomous vehicles are to be widely accepted, we need to ensure their safe operation. For this reason, verification and validation (V\&V) approaches must be developed that are suitable for this domain. Model checking is a formal technique which allows us to exhaustively explore the paths of an abstract model of a system. Using a probabilistic model checker such as PRISM, we may determine properties such as the expected time for a mission, or the probability that a specific mission failure occurs. However, model checking of complex systems is difficult due to the loss of information during abstraction. This is especially so when considering systems such as autonomous vehicles which are subject to external influences. An alternative solution is the use of Monte Carlo simulation to explore the results of a continuous-time model of the system. The main disadvantage of this approach is that the approach is not exhaustive as not all executions of the system are analysed. We are therefore interested in developing a framework for formal verification of autonomous vehicles, using Monte Carlo simulation to inform and validate our symbolic models during the initial stages of development. In this paper, we present a continuous-time model of a quadrotor unmanned aircraft undertaking an autonomous mission. We employ this model in Monte Carlo simulation to obtain specific mission properties which will inform the symbolic models employed in formal verification.
\end{abstract}


\end{frontmatter}

\section{Introduction}

The last few years have seen a huge expansion in the applications and capability of unmanned vehicles. In the military sector, where unmanned vehicles have been in use for decades, efforts are now focussed on increasing the autonomy of such vehicles. That is, increasing their ability to perform tasks with minimal human interference. The commercial sector and hobbyist community are further behind in their development of semi- and fully-autonomous systems. However, autonomous capabilities are just as useful in commercial and civilian applications.

Regardless of the application and their level of autonomy, unmanned vehicles suffer from a number of obstacles which prevent wide scale use. Unmanned aircraft in particular face certification issues related to the dangers posed by a lack of human decision-making in the loop \cite{Webster2011}. Vehicles operating in any uncontrolled environment face a number of unpredictable events and random disturbances, whether internal or sourced from the environment. The ability to appropriately respond to such events and manage such disturbances is a crucial part of any vehicle control system. In the push towards autonomy, this control system must remove the human element from the loop entirely.

Verification is the process of ensuring that a system such as an autonomous vehicle performs as intended. That is, given a specific mission, the vehicle is able to realise the mission goals while responding to any number of events, whether internally or externally derived. Many verification methods exist, ranging from the simple and abstract to the computationally-intensive. Systems such as software packages may be modelled and verified using abstract approaches such as formal verification. More complex systems such as autonomous vehicles are less trivial to abstract. This is due to the constant variation of the vehicle's behaviour, its control system and any environmental factors with time. The presence of random elements in the system adds further complexity.

Due to the intractability of abstracting a complex system, simulation is often used for verification. While still a form of abstraction, continuous- or discrete-time simulation models have the benefit of retaining the dependency of the system and any disturbances on time. Algorithms such as Monte Carlo simulation may then be used to explore the variations in the system due to any stochastic properties such as random events or disturbances.
However, there is the danger that, even after countless iterations, the system has not been exhaustively explored~\cite{Lerda2008}. This differs from formal verification where the analysis is exhaustive, investigating all possible executions of the system, and therefore can find `corner cases' which can be missed with alternative approaches. This is particularly important for autonomous vehicles where guarantees on safety are required. In addition, simulation models are time-consuming to build and run. This is especially so when employed in a Monte Carlo simulation of potentially thousands of iterations.

It is therefore of benefit to re-examine the capabilities of formal verification in ensuring that an autonomous vehicle matches its specification. More specifically, we may consider the use of model checking as a verification method. Model checking involves the construction of a mathematical model usually using a formally specified modelling language that captures the system's behaviour and, to verify that the requirements, usually specified in a temporal logic, are satisfied there is a systematic exploration and analysis of the system model. The drawback of using model checking is the state-space explosion problem, and to perform model checking of complex systems abstraction is often required. PRISM~\cite{Kwiatkowska2011} is a widely used probabilistic model checker which has support for analysing a wide range of quantitative requirements against a number of different quantitative models. PRISM has been applied to the quantitative verification in a wide range of application domains, from wireless communication protocols to power management systems to systems biology.

An autonomous vehicle may be considered a probabilistic system, with the occurrence of random events determined by some probability. Model checking may then explore the impact of such events on the vehicle mission. That is, the paths the vehicle may take to reach a specific finite state, or the ultimate outcome of the mission. However, the symbolic models employed in model checking software are, by their nature, highly abstracted. Some nuances of the system may be lost in abstraction, where a continuous-time simulation model may be able to retain these nuances due to its greater fidelity.

With quantitative model checking of autonomous vehicles being relatively unexplored territory, we wish to determine the differences between the approaches of formal verification and simulation-based verification. To this end, we employ a simple autonomous system, undergoing a basic mission in a controlled environment. We describe the system and environment by a continuous-time model. We then investigate the impact of probabilistic events by evaluating the results of Monte Carlo simulation. It may be noted that more sophisticated simulation techniques are available, for example, guided M-C simulation for counter-example generation \cite{Abbas2013,Sankaranarayanan2012,Nghiem2010}. This method involves attempting to falsify a temporal property by guiding simulation using robustness metrics. Our goal is different. We do not wish to optimise the simulation process itself: the ultimate goal is to produce realistic finite state models for probabilistic verification.

The analysis for this continuous time model are discussed in relation to a concurrent work on a framework for investigate decision-making in autonomous systems, using probabilistic model checking of an abstract model where quantitative data for abstract actions is derived from small-scale simulation models~\cite{Kwiatkowska2011}. More precisely, in this work we have built a abstract model of the same autonomous system using this framework. Here we demonstrate that the Monte Carlo simulation results of the detailed continuous-time model for properties, including the expected time mission completion and the probability of certain mission failures, agree with the bounds obtained using the probabilistic model checker PRISM for the abstract discrete-time model.

The paper is structured as follows. Section~\ref{Sec:Lit_Review} reviews literature relating to formal verification of autonomous vehicles. Section~\ref{Sec:Scenario} describes a case scenario, involving an autonomous quadrotor unmanned aerial vehicle (UAV) with a simple search and retrieve mission. Section~\ref{Sec:Model} describes the continuous-time model of the complete system, including the quadrotor, the environment, the target objects to be retrieved. Section~\ref{Sec:Guidance} describes the guidance system of the quadrotor, which comprises a flight control system (FCS) and finite state machine (FSM), the latter of which provides the vehicle its autonomy. Section~\ref{Sec:Results} presents the results of a single simulation run and those of the Monte Carlo simulation. Section~\ref{Sec:Model_Checking} describes how probabilistic model checking may be applied to a highly-abstracted model of our autonomous mission and compares the results obtained through Monte Carlo simulation and probabilistic model checking. Finally, Section~\ref{Sec:Conclusions} presents the conclusions of the research.

\section{Related Work}\label{Sec:Lit_Review}

A related area of research is that of reachability analysis for the generation of safe plans for automated vehicles~\cite{Althoff2014,Gillula2010,Althoff2015}. Dynamic models of automated vehicles are used to generate a safe set of reachable states (with respect to uncertainties in the movement of vehicles in the system), and thence to generate a safe plan of movement. Our goal is different: we start with a dynamic model of a system (a unmanned aircraft), and use it to (ultimately) derive a faithful finite state probabilistic representation of our system. We then use model checking to examine probabilistic properties of the system (by exploring the smaller, finite state model). Not only can we perform reachability analysis using model checking, but we can analyse our system with respect to a wide range of properties expressed via stochastic logic.

Unmanned vehicles are a complex combination of subsystems in both hardware and software domains. As such, model checking may be used in a variety of investigations in unmanned vehicle verification. Typically, the control system of the vehicle is verified such that its effect on the vehicle behaviour meets the desired specification. However, the purpose of this control system can vary between simple automatic stabilisation and complex decision-making algorithms with multiple subsystems. Here, we discuss existing work in formal verification of control systems for vehicles both manned and unmanned.

Examples of the use of model checking for automatic control systems are  few. This is likely due to the relative simplicity of such a system in comparison to an autonomous or semi-autonomous system. Typically, control systems such as those on aircraft were easily described as simple autopilots. Many more recent flight control systems (FCS) employ autonomy to some degree and, as such, merit the need for verification. A study by Airbus \cite{Bochot2009} investigated the use of model checking for a variety of experiments in verification of embedded systems on A340 and A400M aircraft.

A study by Lerda et al.~\cite{Lerda2008} uses model checking to verify of an embedded control system for a UAV. This investigation uses a continuous-time model, developed in MATLAB, to accurately describe the dynamics of the UAV. Model checking is then used to verify the control system. The study highlights the benefits of both simulation and model checking. Simulation can be cumbersome and is often prohibitively time-consuming when exploring the many aspects of a system's behaviour. Conversely, model checking of a continuous-time system is difficult, as the state-space of a continuous system is infinite in size. The use of hybrid automata is one solution to this problem \cite{Henzinger1996,Chutinan2000,Kowalewski1999}, however the formal verification techniques which are applicable to such a model are computationally expensive. Additionally, the complexity of systems which can be analysed using this approach is low. This precludes application to autonomous systems, which are complex by their very nature.

Many instances of model checking for autonomous vehicles involve swarms. Typically, the use of a group of vehicles to perform some task is investigated. Model checking is then used to identify the routes that may be taken to accomplish this task. Examples include UAVs in a pursuer-evader \cite{Bohn2007} and foraging \cite{Konur2012,Liu2007,Mikael2012} scenarios. Many of these investigations describe the behaviour of each entity in the swarm as a finite state machine (FSM). In a simple model, the number of entities and their states determine the size of the model's state-space. The dynamics of the vehicles are considered in some cases through the use of simulation \cite{Liu2007,Mikael2012}. However, in these instances, the vehicles in question are relatively simple in their mechanics.

Other applications of model checking to autonomous vehicles focus on verifying the interaction between agents in co-operative missions \cite{Humphrey2013}. More recent work focusses on the verification of autonomous behaviours with a view to obtaining certification of UAVs for widespread commercial use \cite{Webster2011,Choi2012}.

\section{Scenario}
\label{Sec:Scenario}

\noindent
We employ a simple scenario as a case study. A quadrotor UAV is in operation inside a small, constrained environment. This environment is based on the University of Glasgow's Micro Air Systems Technologies (MAST) Laboratory, a $4 \times 7 \times 3$ \si{\metre} cuboidal flight space with an Optitrack\footnote{Natural Point, Inc. https://www.optitrack.com}  motion capture system for tracking UAVs. Three inanimate target objects, each of unique colour and shape, are located somewhere on the floor of the lab. The UAV is programmed such that it should take off from a specified landing site and following a predetermined series of waypoints, while searching for these objects. Upon finding a target, the quadrotor retrieves it from the floor, transports it to a predetermined drop site and deposits it. It then continues following the waypoints until all of the targets have been relocated or it reaches the final waypoint. It then returns to base, having either succeeded in its mission by relocating all targets or failed by having missed at least one.

Stochastic behaviour is introduced into the scenario through the inclusion of faults, the occurrence of which are based on a uniform probability distribution. Certain system properties are defined similarly, such as the initial locations of the targets, the landing site of the quadrotor and the drop site. These are described in greater detail in Section~\ref{Sec:Stochastic_Vars}.

These random properties can propagate through the model, ultimately affecting the outcome of the mission. The success of the mission is thus a function of the target, drop site and landing site locations and the occurrence of faults. This function is then defined by the model of the autonomous system. This model is a complex combination of hardware and software subsystems.

\section{Mathematical Model}
\label{Sec:Model}

The mathematical model describes a multi-agent system and is implemented using an object-oriented programming (OOP) approach. The quadrotor may be considered as either an \textit{active} or \textit{cognitive} agent \cite{Kubera2010}. The targets can be considered as \textit{passive} agents \cite{Kubera2010}, in that they have their own dynamics but no goals or reactive behaviours. They may alternatively be considered \textit{objects} \cite{Frank2001}, in that they may be modified by the UAV agent. We choose the former definition for the targets, such that we may denote a generic agent $\ag$ and use it to specify properties or behaviours common to both UAV and targets. All agents exist with an environment, which corresponds to the MAST Laboratory in this instance.

The UAV agent has more complex behaviours than the targets. Its guidance software contains both a flight control system and decision-making algorithms. It is these decision-making algorithms which allow the UAV to perform its mission with no interference from a human operator.

In this section, we describe the mathematical model of the multi-agent system, comprising the UAV, targets and environment. First, we define the overall structure of the model. That is, how the agents interact with one another and the environment. Then, we define the frames of reference used to describe the behaviours of the agents in a local body-fixed axes system. We then provide the framework for the geometrical models of the simulation, including the environment. Models of the quadrotor hardware are defined, including the vehicle dynamics and sensor systems. Target dynamics are described similarly. Finally, the stochastic variables used to introduce probabilities into the model are detailed.

\subsection{Structure}
\begin{figure}
	\centering
	\includegraphics[scale=0.8]{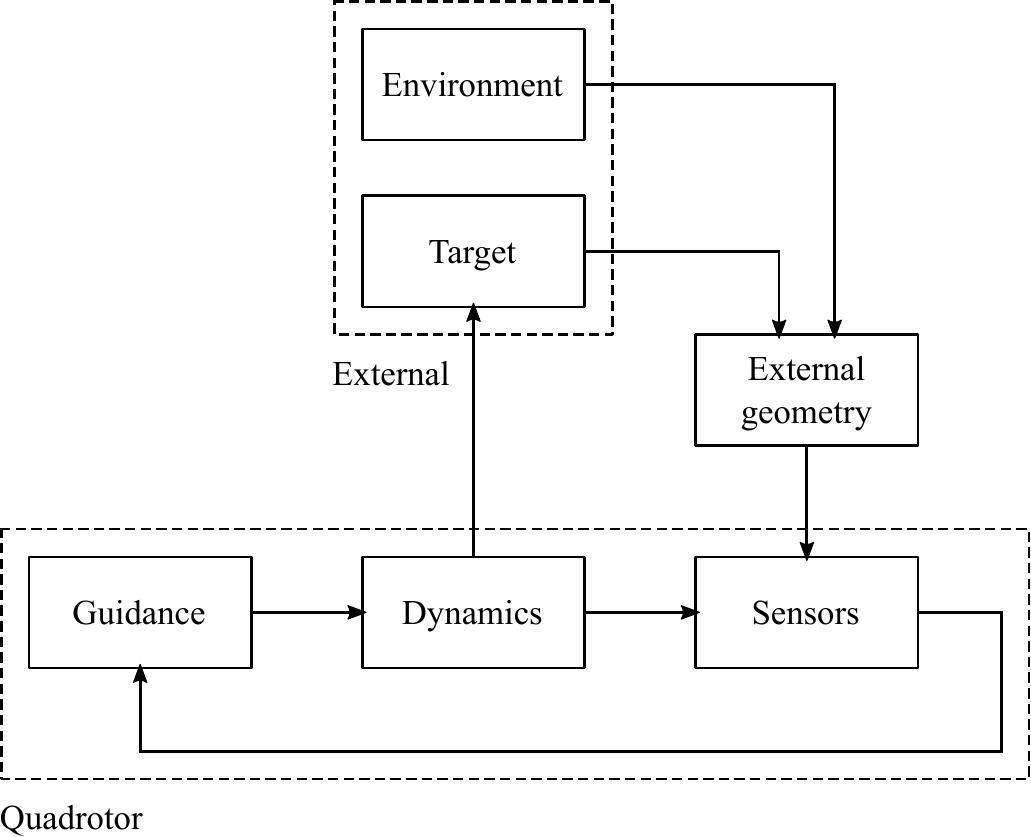}
	\caption{Block diagram of multi-agent system.}
	\label{Fig:Block_Diagram}
\end{figure}

Figure~\ref{Fig:Block_Diagram} shows a simplified diagram of the multi-agent system. The UAV's sensor suite senses both the internal states of the quadrotor and the geometry of the targets and environment.  The sensor measurements are utilised by the guidance system of the UAV, which then provides control inputs to the UAV's rotors. The guidance system comprises several subsystems which allow the UAV to interpret the sensor feedback, decide on a course of action and drive the vehicle towards that action. This is described in greater detail in Section~\ref{Sec:Guidance}. When the UAV grasps a target, the target behaviour then becomes dependent on the response of the UAV. The dynamics of the UAV hence directly impact the target response during grasping. The environment is static and hence unaffected by the behaviour of the agents.

\subsection{Frames of Reference}
The positions and orientations of agents are described in an inertially-fixed \textit{World} frame of reference, denoted $\Wframe$. This frame is defined such that $z^\Wframe$ is in the direction of the local gravity vector, while $x^\Wframe$ and $y^\Wframe$ are parallel to the boundaries of the environment and satisfy the rule $\hat{\vc{x}} = \hat{\vc{y}} \times \hat{\vc{z}}$. The position of an agent $\ag$ is described in $\Wframe$ by $\vc{r}_\ag \in \mathbb{R}^3$, where it is assumed that the agent body is rigid.

We can define a second frame of reference $\Aframe$, which is fixed on any agent $\ag$ and has origin at $\vc{r}_\ag$, which corresponds to the agent centre of mass. The transformation from $\Wframe$ to $\Aframe$ is described by the direction cosine matrix
\begin{equation}
\begin{aligned}
	\vc{R}_\Wframe^\Aframe &= \begin{bmatrix}
		\ctheta \cpsi	& \ctheta \spsi	& -\stheta \\
		\sphi \stheta \cpsi - \cphi \spsi	& \sphi s_\theta \spsi + \cphi \cpsi	& \sphi \ctheta \\
		\cphi \stheta \cpsi + \sphi \spsi	& \cphi \stheta \spsi - \sphi \cpsi	& \cphi \ctheta
	\end{bmatrix}
\end{aligned}
\label{Eq:Generic_DCM}
\end{equation}

Thus, any position $\vc{q}^\Aframe$ in $\Aframe$ may be described in $\Wframe$ by evaluating the relationship $\vc{q}^\Wframe = \vc{R}_\Aframe^\Wframe \vc{q}^\Aframe$. The reverse transformation $\vc{q}^\Aframe = \vc{R}_\Wframe^\Aframe \vc{q}^\Wframe$, where $\vc{R}_\Aframe^\Wframe = ( \vc{R}_\Wframe^\Aframe )^T$ similarly applies.

The orientation is alternatively described by the vector $\vc{\eta}_\ag = [\phi,\theta,\psi]^T$. The rotation of $\Aframe$ is described by the angular velocity vector $\vc{\omega}_\ag \in \mathbb{R}^3$. The rate of change of orientation $\dot{\vc{\eta}}_\ag$ with respect to $\Wframe$ is then described by the relationship
\begin{equation}
\begin{aligned}
	\dot{\vc{\eta}}_\ag &= \vc{J}_\Aframe^{-1} \vc{\omega}_\ag \\
	\text{where}\quad\vc{J}_\Aframe^{-1} &= \begin{bmatrix}
		1	& \sin\phi\tan\theta	& \cos\phi\tan\theta \\
		0	& \cos\phi				& -\sin\phi \\
		0	& \sin\phi\sec\theta	& \cos\phi\sec\theta
	\end{bmatrix}
\end{aligned}
\label{Eq:Omega_Euler}
\end{equation}

We can then state that any rigid body agent $\ag$ in $\Wframe$ has position $\vc{r}_\ag(t) \in \mathbb{R}^3$ and orientation (or attitude) $\vc{\eta}_\ag(t) \in \mathbb{R}^3$. The position and attitude responses of a given agent are determined by the dynamics of that agent.

\subsection{Geometry}
\label{Sec:Geometry}
The geometry of the model is defined as a series of polygons with associated faces, vertices and colours. In general, any geometrical object is composed of a series of faces. Each face has an associated colour. The shape, position and orientation of a face in a given reference frame is determined by the number and positions of the vertices associated with that face. The primary use of the geometrical models is to visualise the simulation through animation. However, the models are also used by the UAV's camera sensor model to sense the environment and target agents.

The geometry of any agent $\ag$ is composed of $m$ vertices, where each vertex $i \in \{1,2,\ldots m\}$, with position $\vc{v}_i \in \mathbb{R}^3$, is associated with one or more faces. If the geometry has $n$ faces, a single face $j~\in~\{1,2,\ldots n\}$ of $p$ vertices is then defined by the vector $\vc{f}_j$. Each face $j$ has an associated colour, defined by the vector $\vc{C}_j \in [0, 1]^3$.

Colour detection is used to identify targets. If we state that each colour $\vc{C}_j$ corresponds to a specific face $j$, we may obtain the indices $i$ of the vertices associated with that face and colour from the elements of $\vc{f}_j$. We may then ignore vertices associated with colours outside of a chosen range. The visibility of remaining vertices to the camera sensor is evaluated by considering their positions $\vc{v}_i$ in camera image space. This is described in greater detail in Section~\ref{Sec:Object_Tracking}.

The position $\vc{v}_i^\Aframe$ of any agent vertex $v_i$ is fixed in the local frame of reference $\Aframe$. It may be described in $\Wframe$ by considering the transformation
\begin{equation}
	\vc{v}_i^\Wframe = \vc{R}_\Aframe^\Wframe \vc{v}_i^\Aframe + \vc{r}_\ag
\end{equation}
where $\vc{r}_\ag$ is the position of the origin of $\Aframe$ in $\Wframe$.

\subsection{Environment}
The environment is defined as a simple arena which imposes limits on the freedom of the agents. The World frame $\Wframe$ is defined as beng at the geometric centre of the floor of the environment. The $z^\Wframe$-axis is defined in the direction of the local gravity vector. The position of any object or vertex in $\Wframe$ can be described by $\vc{r} = [x,y,z]^T$. The environment then imposes the limits $x \in [-2,2]$, $y \in [-3.5,3.5]$, $z \in [-3,0]$.

The geometry of the environment is limited to defining the location of the \textit{drop site} $\vc{r}_\ds$, where the collected targets are deposited. We may define the \textit{drop zone} as the interior of a circle centred at $ \vc{r}_\ds $ with radius $R_\ds$. The UAV perceives targets inside the drop zone, but otherwise ignores them.

\subsection{Quadrotor Dynamics}
A quadrotor is used as the UAV platform, owing to its versatility and mechanical simplicity. A quadrotor typically consists of four fixed-pitch rotors spaced equally around the centre of mass of the vehicle airframe. The rotors produce both lifting force and, by creating thrust differentials between opposing pairs of rotors, enable control over vehicle attitude. The behaviour of the quadrotor is detailed in  \cite{Bouabdallah2004,Voos2009,Ireland2014}.

For the search and retrieve scenario, we employ a standard quadrotor UAV with some additional components. The quadrotor is considered a rigid body, with the primary force and moment contribution from the four rotors and the vehicle weight. To facilitate its interaction with any target, we describe the dynamics of a rigidly-mounted grasping arm separately. We then introduce a simple gimbal-stabilised platform for the UAV's camera sensor, and a simple power consumption model.

\subsubsection{Rigid-Body Dynamics}
We define a quadrotor body-fixed frame $\Qframe$ which has origin at the vehicle centre of mass, $x^\Qframe$ in the nominal forward direction, $y^\Qframe$ in the nominal starboard direction and $z^\Qframe$ such that the condition $\hat{\vc{z}} = \hat{\vc{x}} \times \hat{\vc{y}}$ is satisfied. The position of the vehicle in $\Wframe$ is denoted $\vc{r}_\qud$ and its attitude by $\vc{\eta}_\qud$.

\begin{figure}
	\centering
	\includegraphics[scale=0.9]{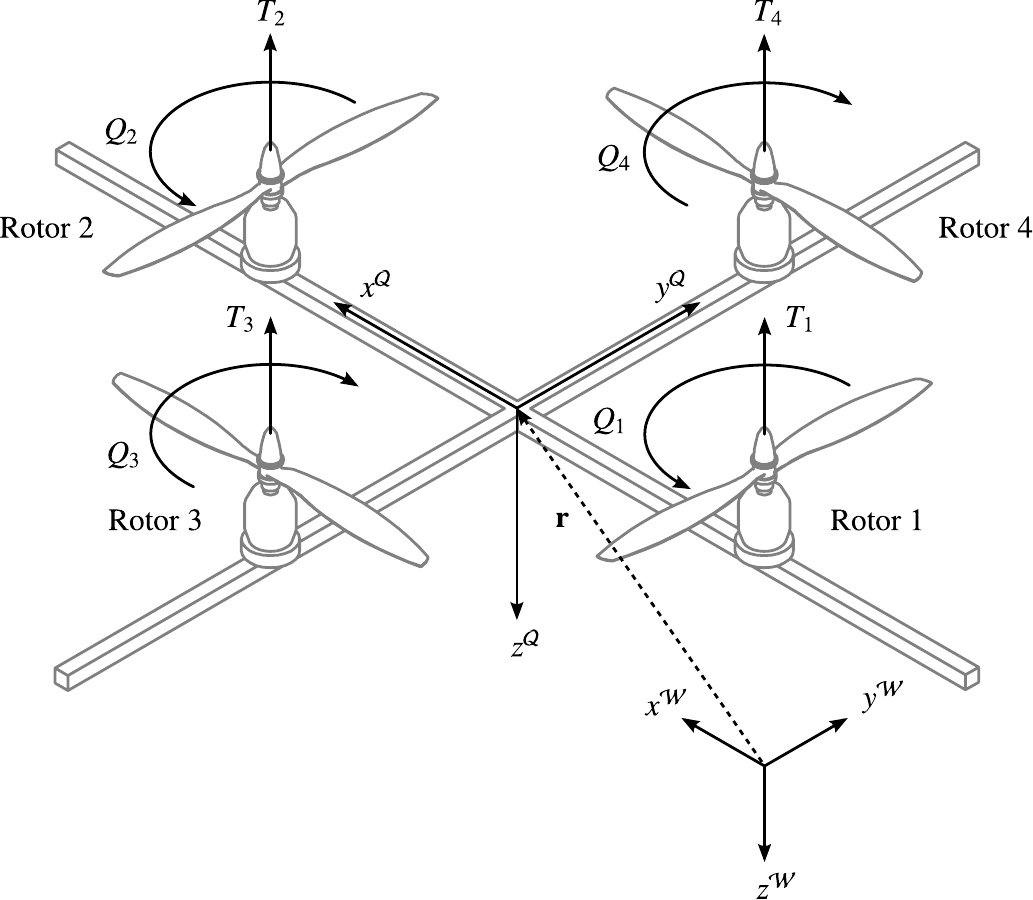}
	\caption{Quadrotor body-fixed frame definition and rotor arrangement.}
	\label{Fig:Quad_Dynamics}
\end{figure}

Next, we employ a standard quadrotor dynamic model, which assumes a rigid-body response with six degrees of freedom \cite{Voos2009,Tayebi2006}. The rotor dynamics are assumed to be sufficiently faster than the closed-loop body dynamics \cite{Voos2009}, such that they may be considered instantaneous. A linear relationship between thrust and input PWM signal is assumed \cite{Ireland2014}. The rotor arrangement shown in Figure~\ref{Fig:Quad_Dynamics} is employed.

We make the addition of a simple mass-spring-damper model to represent the vehicle interaction with the floor of the environment. Any collision with the floor is thus described as a highly-damped spring. This ``floor force'' applies only when the floor plane intersects the rigid geometry of the UAV. The geometry of the UAV is thus characterised by an \textit{effective radius} $R_\qud$ around the centre of mass. When the floor plane intersects with this sphere of radius $R_\qud$ about $\vc{r}_\qud = [x,y,z]^T$, a force is exerted. Otherwise, no force is exerted.

This force, denoted $\vc{F}_f$, is assumed to produce no moment and act normal to the floor plane only, that is in the $z^\Wframe$ direction. We thus describe it by the conditional relationship
\begin{equation}
\begin{aligned}
	\vc{F}_f &= F_f \vc{R}_\Wframe^\Qframe \hat{\vc{z}} \\
	\text{where } F_f &= \begin{cases}
	-k_\qud(R_\qud + z_\qud) - c_\qud \dot{z} & \text{if } z_\qud \geq -R_\qud \\
	0 & \text{if } z_\qud < -R_\qud
	\end{cases}
\end{aligned}
\label{Eq:Floor_Model_Quad}
\end{equation}
where $k_\qud$ is the spring stiffness and $c_\qud$ the damping coefficient. These parameters determine the ``spring'' of the floor. That is, how much the UAV bounces upon collision with the floor.

The quadrotor dynamics are thus described by the 12-state non-linear model
\begin{equation}
\begin{aligned}
	\ddot{\vc{r}}_\qud &= \left( g + \frac{F_f}{m_\qud} \right) \hat{\vc{z}} - \frac{K_T}{m_\qud} \sum_{k=1}^4 u_k\, \vc{R}_\Qframe^\Wframe \hat{\vc{z}} \\
	\dot{\vc{\eta}}_\qud &= \vc{J}^{-1}_\Qframe \vc{\omega}_\qud \\
	\dot{\vc{\omega}}_\qud &= \vc{I}_\qud^{-1} \left(\begin{bmatrix}
		LK_T (u_2 - u_1) \\
		LK_T (u_3 - u_4) \\
		K_Q(-u_1 - u_2 + u_3 + u_4)
	\end{bmatrix} - \vc{\omega}_\qud \times \vc{I}_\qud \vc{\omega} _\qud \right)
\end{aligned}
\label{Eq:Quad_Dynamics}
\end{equation}
where $m_\qud$ is the quadrotor mass and $\vc{I}$ its inertia tensor, $g$ is the acceleration due to gravity, $K_T$ and $K_Q$ are rotor gains, $L$ is the rotor moment arm and $F_f$ is subject to the conditions imposed by Equation \eqref{Eq:Floor_Model_Quad}. The properties of the quadrotor model are detailed in Table~\ref{Tab:Sim_Properties}. These are obtained from system identification of the Quanser Qball-X4 quadrotor, as described in \cite{Ireland2014}.

\subsubsection{Grasper Arm Dynamics}
When a target is tethered to the quadrotor, its dynamics are coupled to those of the grasper arm. The dynamics of the arm are thus considered separately from the UAV's rigid-body response. The grasper has fixed position $\vc{r}_{\grb/\qud}^\Qframe \in \mathbb{R}^3$ in the rotating frame $\Qframe$. It therefore has the inertial position
\begin{align*}
	\vc{r}_\grb = \vc{r}_\qud + \vc{R}_\Qframe^\Wframe \vc{r}_{\grb/\qud}^\Qframe
\end{align*}
with the second derivative
\begin{align}
	\ddot{\vc{r}}_\grb &= \ddot{\vc{r}}_\qud + \vc{R}_\Qframe^\Wframe  \left[ \vc{\omega}_\qud \times \left( \vc{\omega}_\qud \times \vc{r}_{\grb/\qud}^\Qframe \right) + \dot{\vc{\omega}}_\qud \times \vc{r}_{\grb/\qud}^\Qframe \right]
\label{Eq:Grasper_Dynamics}
\end{align}
where translational and rotational accelerations, $\ddot{\vc{r}}_\qud$ and $\dot{\vc{\omega}}_\qud$ respectively, are given by Equation \eqref{Eq:Quad_Dynamics}. The rotational dynamics of the grasper are identical to that of the quadrotor.

\subsubsection{Gimbal Dynamics}
The camera sensor is mounted on a gimbal which stabilises it during rolling and pitching motion. When the UAV is level, the principal axis of the camera is in the direction of $z^\Wframe$. We define $x^\Cframe$ in this direction, while $y^\Cframe$ and $z^\Cframe$ are in the directions of the camera image's horizontal and vertical directions, respectively.

When the UAV rolls or pitches the gimbal platform rotates to counter this displacement. The roll $\phi_g$ and pitch $\theta_g$ dynamics of the gimbal platform may be described by the first-order system
\begin{equation}
	\begin{bmatrix}
		\dot{\phi}_g \\ \dot{\theta}_g
	\end{bmatrix} = -\frac{1}{\tau_g}\begin{bmatrix}
		\phi + \phi_g \\
		\theta + \theta_g
	\end{bmatrix}
\end{equation}
where $\tau_g$ is the time constant of the gimbal rotational response.

It is assumed that the camera centre is coincident with both the centre of rotation of the gimbal platform and the origin of $\Cframe$. The relationship between $\Cframe$ and $\Qframe$ is then described by the direction cosine matrix
\begin{equation}
\begin{aligned}
	\vc{R}_\Qframe^\Cframe &= \begin{bmatrix}
		\sin\theta_g & 0 & \cos\theta_g \\
		0 & 1 & 0 \\
		-\cos\theta_g & 0 & \sin\theta_g
	\end{bmatrix} \begin{bmatrix}
		1 & 0 & 0 \\
		0 & \cos\phi_g & \sin\phi_g \\
		0 & -\sin\phi_g & \cos\phi_g
	\end{bmatrix} \\
	&= \begin{bmatrix}
		\sin\theta_g & -\sin\phi_g\cos\theta_g & \cos\phi_g\cos\theta_g \\
		0 & \cos\phi_g & \sin\phi_g \\
		-\cos\theta_g & -\sin\phi_g\sin\theta_g & \cos\phi_g\sin\theta_g
	\end{bmatrix} \\
	\vc{R}_\Cframe^\Qframe &= \left( \vc{R}_\Qframe^\Cframe \right)^T
\end{aligned}
\end{equation}

\subsubsection{Battery Model}
A battery model is employed to allow us to introduce the complication of the UAV periodically returning to base to recharge. As the battery discharges, the voltage supplied to the UAV reduces. This reduction is described by the first-order model
\begin{equation}
	\dot{V} = \begin{cases}
		v_c & \text{if charging and } V < V_{\max} \\
		v_d & \text{if not charging} \\
		0 & \text{if } V \geq V_{\max}
	\end{cases}
\end{equation}
where $v_c > 0$ is the charging rate, which applies when the UAV is idle at the landing site, $v_d < 0$ is the discharge rate, which applies when the UAV is pursuing its mission. The voltage level has the upper limit $V \leq V_{\max}$, above which it will not charge. The discharge rate is approximated as a constant. In reality, the battery drain is affected by the power requirements of the rotors and other hardware components.

\subsubsection{Stochastic Properties of the Quadrotor}
We have previously described a simple deterministic quadrotor model. The model has enough detail that it may realistically describe a quadrotor which is capable of sitting idle on the ground, flying and grasping objects. In reality, the quadrotor is subject to a number of stochastic events. Through their effect on the system behaviour, these events can impact the particulars of the mission, including its outcome. Here, we discuss some of these events and, if relevant, how we have chosen to incorporate these into our model.

First, we may consider the effect of external disturbances. Such disturbances are products of the local environment and are governed by a system far greater in complexity than the quadrotor itself. Thus, we may consider them to be random. These disturbances may manifest as forces or moments acting on the vehicle body. They may also impact the performance of the rotors by changing the local airflow around the rotor disk. A robust control system is typically employed to reduce the impact of such disturbances. In reality, even with robust control, these disturbances may have the effect of changing the path the UAV takes, or the time it takes to perform a manoeuvre.

The quadrotor system may also be subject to internal disturbances. These may manifest as a change in a parameter which is expected to be constant. For example, it is assumed that the rotor thrust gain $ K_T $ does not change. However, the thrust gain can be altered by a number of phenomena, including varying battery level and local airflow around the rotor, as indicated above. One or more rotors may also fail entirely, due to a fault in any of the components between autopilot and rotor, such as the motor or propeller. Regardless of where the fault occurs, the effect is a complete loss of thrust and torque from that rotor.

The occurrence such an \textit{actuator fault} is based on a large number of variables. We abstract the probability of such a fault occurring by using a geometric distribution assuming that, with the fixed period $ T_a $, the probability of an actuator fault equals $ P_a: \mathbb{R}_{\geq 0} \rightarrow [0,1]$ (see Table~\ref{Tab:Sim_Properties}). We model the actuator fault as a complete loss of thrust in a single rotor, that is $ K_{Tk} = 0 $ for a random rotor $ k \in \{1,2,3,4\} $. For example, if a fault occurs in the rear rotor $ k=1 $, the maximum propulsive force is reduced by a quarter and the $u_1$ terms in Equation \eqref{Eq:Quad_Dynamics} may be neglected.

While control strategies are available to deal with an actuator fault \cite{Mueller2014}, the reduced capability of the vehicle typically results in a controlled emergency landing, if not a crash landing. In Section~\ref{Sec:Control_Modules}, we introduce a control algorithm specifically for emergency landings after actuator loss.

A fault in the UAV's grasping mechanism may result in the mechanism becoming stuck in its current state or it may release any tethered objects. Since the former may result in a target being permanently tethered to the UAV and thus preventing mission success, we limit our model to include the later case only. It is assumed that some fault occurs which causes the grasper mechanism to deactivate. This has no effect when no target is being grasped. However, if a target is held by the mechanism, it is released and the UAV must retrieve it again to continue the mission. Just as we define the probability of an actuator fault occurring, we may assume that the probability of a \textit{grasper fault} occurring within a fixed time period $T_g$ equals $P_g: \mathbb{R}_{\geq 0} \rightarrow [0,1]$. The impact on the UAV behaviour due to a grasper fault is detailed in Section~\ref{Sec:Guidance_Modes}.

Finally, we randomly define the pose (position and heading) of the UAV at the beginning of the mission. This has the effect of altering the behaviour of the UAV as the predicates which trigger transitions from mode to another may occur at different times and under different circumstances. That the target locations and drop site are also randomly defined further contributes to the uncertainty in the system.

\subsection{Quadrotor Sensors}
The quadrotor employs a standard sensor suite to measure system states and sense the environment. A motion capture system allows direct measurement of vehicle position and attitude. An inertial measurement unit (IMU), containing tri-axial accelerometers and gyroscopes, measures accelerations and angular rates in $\Qframe$. A camera sensor mounted on the underside of the vehicle is used to visually sense the geometry of the environment and targets. This section describes the models of the sensor behaviour. Interpretation and exploitation of sensor outputs is described in Section~\ref{Sec:Guidance}.

\subsubsection{Motion Capture System}
The position and attitude of the UAV in $\Wframe$ are measured by the motion capture system. The model is based on the MAST Laboratory's Optitrack system . Neglecting measurement noise, we may assume the motion capture system measures both position and attitude with zero error, giving
\begin{align}
	\hat{\vc{o}}_r &= \vc{r}_\qud, & \hat{\vc{o}}_\eta &= \vc{\eta}_\qud
\end{align}

\subsubsection{Inertial Measurement Unit}
The angular rates of the UAV in $\Qframe$ are measured by the onboard IMU. Again neglecting any noise or bias errors, we describe the gyroscope measurement by
\begin{equation}
	\hat{\vc{g}} = \vc{\omega}_\qud
\end{equation}

\subsubsection{Camera Sensor}
\label{Sec:Camera_Sensor}
The camera sensor provides an image of any unobscured object within its field of view. In reality, the camera captures an image which is then used in an object detection algorithm. Examples of such algorithms range from simple colour or intensity detection to more complex shape or pattern recognition. The camera model is described in such a way that it enables the process of colour detection to be emulated. This process is further detailed in Section~\ref{Sec:Object_Tracking}.

\begin{figure}
	\centering
	\includegraphics[scale=0.9]{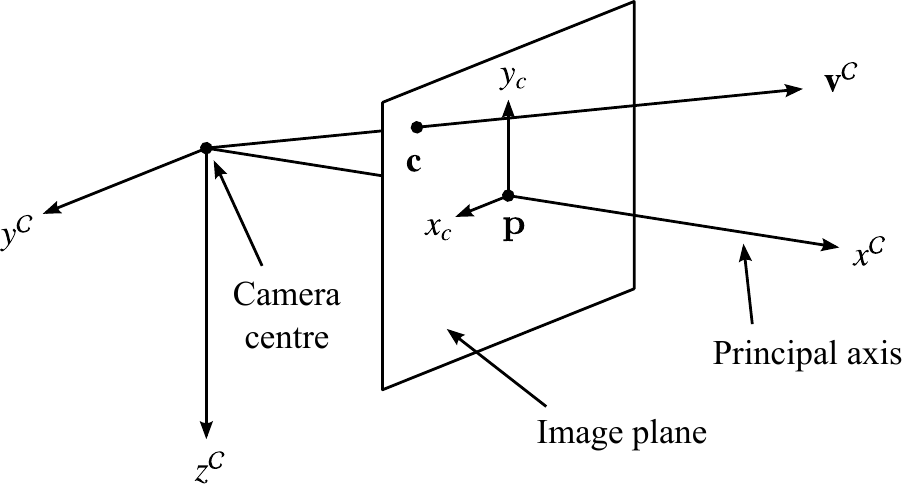}
	\caption{Geometry of a pinhole camera. The coordinates of a point with position $\vc{r}$ in Euclidean 3-space are mapped to 2-space be considering the interaction of the point with the image plane \cite{Hartley2003}.}
	\label{Fig:Camera_Model}
\end{figure}

As detailed in Section~\ref{Sec:Geometry}, any vertex $i$ of a geometrical model has position $\vc{v}_i^\Wframe$ in $\Wframe$. The camera operates by projecting geometry within the its field of view onto an \textit{image plane}, demonstrated in Figure~\ref{Fig:Camera_Model}. In order to achieve this projection, the vertex position must be determined in a frame $\Cframe$, fixed on the camera body.

The camera has fixed position $\vc{r}_{\cam/\qud}^\Qframe$ and orientation $\vc{R}_\Cframe^\Qframe$ in $\Qframe$. Any vertex $\vc{v}^\Wframe$ in $\Wframe$ thus has position, relative to the origin of $\Cframe$, of
\begin{equation}
	\vc{v}^\Cframe = \vc{R}_\Qframe^\Cframe \left[ \vc{R}_\Wframe^\Qframe \left( \vc{v} - \vc{r}_\Qframe \right) + \vc{r}_{\cam/\qud}^\Qframe  \right]
\end{equation}

The vertex is now defined relative to the camera and centre. If $\vc{v}^\Cframe = [v_x,v_y,v_z]^T$, then the projection of $\vc{v}^\Cframe$ onto the image plane is described by the coordinates
\begin{align}
	x_c &= f \frac{v_y}{v_x}, &
	y_c &= -f \frac{v_z}{v_x}
\end{align}
where $f$ is the focal length of the camera.

The camera has horizontal field of view $\lambda$ and aspect ratio $A$. For any given vertex $\vc{c} = [x_c,y_c]^T$ to be visible to the camera sensor, it must be within the field of view. The conditions
\begin{equation}
\begin{aligned}
	-f \tan \lambda  &\leq x_c \leq f \tan \lambda \\
	-\frac{f}{A} \tan \lambda  &\leq y_c \leq \frac{f}{A} \tan \lambda
\end{aligned}
\label{Eq:Camera_Limits}
\end{equation}
must therefore be satisfied.

\subsection{Target Model}
While static for the majority of the mission, each target is given dynamic behaviours in order to facilitate its interaction with the UAV. In isolation, the behaviour of a target is purely deterministic. However, any interactions with the UAV are subject to the influence of stochastic and probabilistic behaviours.

\begin{figure}
	\centering
	\includegraphics[scale=0.5]{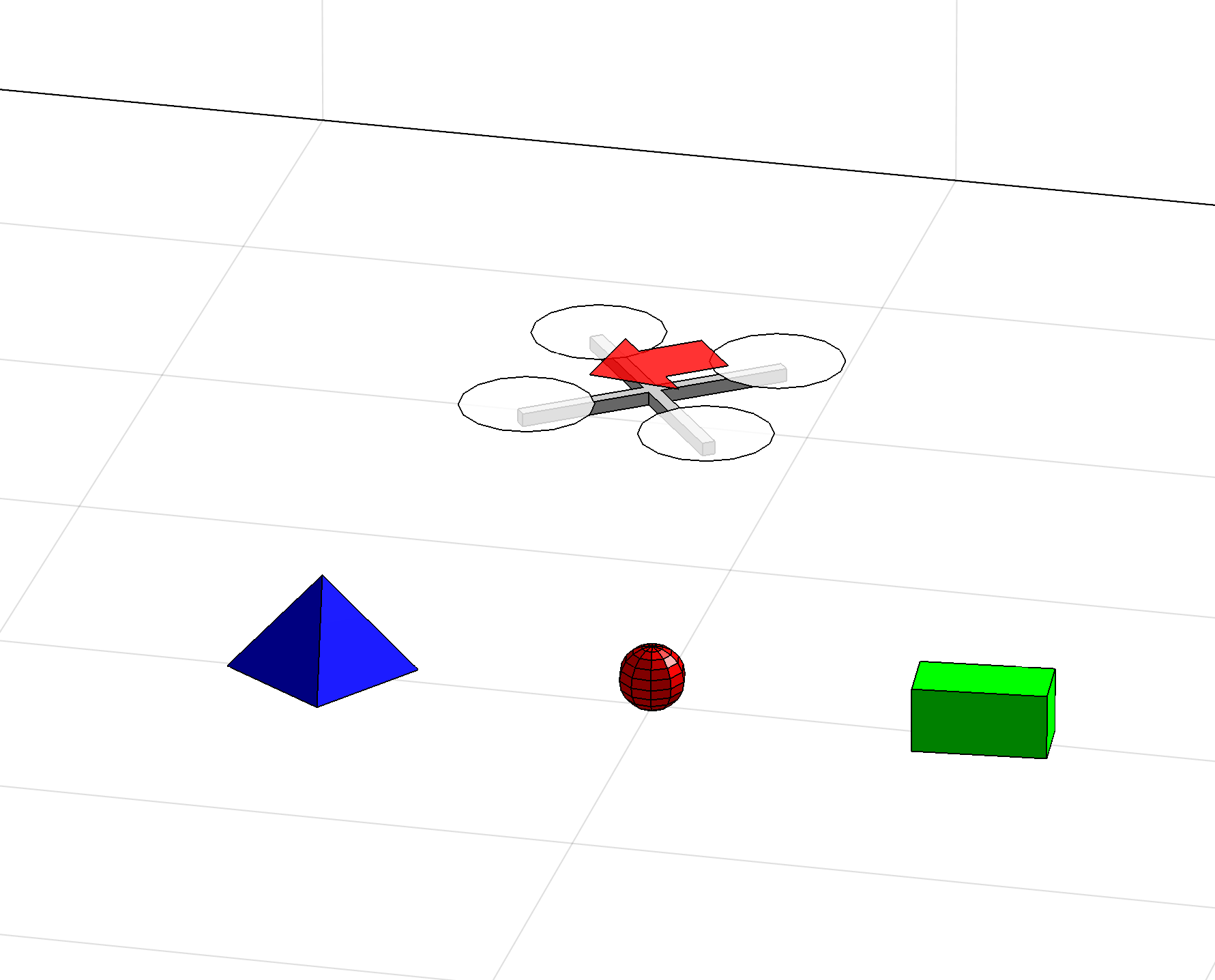}
	\caption{Target geometries and colours as rendered by simulation visualisation.}
	\label{Fig:Target_Colours}
\end{figure}

While each target has the same dynamic behaviour, we consider three distinct shapes and colours. This allows the UAV to discriminate between targets based on their colour and, in the event of more complex object detection being required, their shape. The targets may be seen in Figure~\ref{Fig:Target_Colours}. We model the three targets as a blue pyramid, red sphere and green cuboid, respectively.

\subsubsection{Rigid-Body Dynamics}
We may consider each target to be a rigid body of mass $m_\trg$. A target body-fixed frame $\Tframe$ is defined with origin at the centre of mass of the target and $\vc{R}_\Tframe^\Wframe = \vc{I}_3$ before grasping occurs, where $\vc{I}_3$ is the $3\times 3$ identity matrix.

When tethered to the UAV during grasping, the dynamics of the target are identical to those of the UAV grasping arm, and are therefore given by
\begin{equation}
\begin{aligned}
	\ddot{\vc{r}}_\trg &= \ddot{\vc{r}}_\grb \\
	\dot{\vc{\eta}}_\trg &= \vc{J}_\Tframe^{-1} \vc{\omega}_\trg \\
	\dot{\vc{\omega}}_\trg &= \dot{\vc{\omega}}_\qud
\end{aligned}
\label{Eq:Target_Dynamics_Tethered}
\end{equation}
where $\ddot{\vc{r}}_\grb$ and $\dot{\vc{\omega}}_\qud$ are defined by Equations \eqref{Eq:Grasper_Dynamics} and \eqref{Eq:Quad_Dynamics}, respectively.

When not tethered to the UAV, the target dynamics are driven by a simple gravitational force and floor model
\begin{equation}
\begin{aligned}
	\ddot{\vc{r}}_\trg &= g \hat{\vc{z}} + \frac{1}{m_\trg} \vc{F}_f \\
	\dot{\vc{\eta}}_\trg &= \vc{J}_\Tframe^{-1} \vc{\omega}_\trg \\
	\dot{\vc{\omega}}_\trg &= \vc{0}
\end{aligned}
\label{Eq:Target_Dynamics_Untethered}
\end{equation}
where $\vc{F}_f$ is the force exerted by the floor on the target. This now includes a frictional component in the horizontal plane and is given by
\begin{align}
	\vc{F}_f = \begin{cases}
		\vc{R}_\Wframe^\Tframe \left( -k_\trg \left( R_\trg + z_\trg \right) \hat{\vc{z}} - c_\trg \dot{\vc{r}}_\trg \right)  & \text{if } z_\trg \geq -R_\trg \\
		\vc{0}_{3\times1} & \text{if } z_\trg < -R_\trg
	\end{cases}
\end{align}
where $R_\trg$ is the effective radius of the target and $k_\trg$, $c_\trg$ represent the spring stiffness and damping of the floor-target interaction. This floor model allows us to describe the behaviour of the target as it impacts the floor and comes to rest.

\subsubsection{Stochastic Properties of the Target}
The internal behaviour of the target as described by Equation \eqref{Eq:Target_Dynamics_Untethered} is purely deterministic. However, when grasped by the UAV, its behaviour is determined by that of the UAV (Equation \eqref{Eq:Target_Dynamics_Tethered}). In this state, any stochastic behaviours demonstrated by the UAV will similarly impact the target.

The actuator fault described previously has the effect of prematurely ending the mission, thus its effect on the target is largely inconsequential. Conversely, a fault in the grasping mechanism has significant impact on both the UAV and target.

When a fault occurs in the UAV's grasping mechanism, any target currently tethered to the UAV is released. The target then follows a ballistic trajectory before colliding with the ground. In this phase, the state transition of the target is purely deterministic (Equation~\ref{Eq:Target_Dynamics_Untethered}), although the state vector at the beginning of the trajectory is determined by the state of the UAV at release, which is stochastic. The eventual resting place of the target after being dropped is thus similarly random.

Similarly, the initial pose of each target is randomly defined. This impacts the behaviour of the UAV and thus the time and location at which the target is picked up by the UAV and deposited in the drop zone.

\subsection{Summary of Modelled Stochastic Properties}
\label{Sec:Stochastic_Vars}
As detailed previously, both the UAV and target agents are subject to non-deterministic behaviours. These may be caused by randomly-define properties or by events which have a certain probability of occurring.

\subsubsection{Random Properties}
As described in the models for each agent type, the initial positions of the UAV and targets on the $x^\Wframe$-$y^\Wframe$ plane are defined at random. The headings $\psi$ are similarly specified. If an agent $\ag$ has position $\vc{r}_{\ag,0} = [x_0,y_0,z_0]^T$ at time $t= 0$, then the position in the horizontal plane is defined randomly in the ranges $x_0 \in [-1.9, 1.9]$ and $y_0 \in [-3.4, 3.4]$. The initial height is defined $z_0 = -R_\ag$, where $R_\ag$ is the effective radius of the agent. The drop site location $\vc{r}_\ds = [x_\ds,y_\ds,z_\ds]^T$ is similarly defined, with $x_\ds \in [-1.5, 1.5]$, $y_\ds \in [-3, 3]$ and $z_\ds = 0$. The random properties are generated using a uniform distribution.

\subsubsection{Probabilistic Events}
\label{Sec:Probalistic_Events}
We have described the effects of actuator and grasper mechanism faults on the UAV system. We may implement these faults in our simulation by evaluating the probability of their occurrence in a given time step $ dt $.

An actuator fault may occur at any time. If the UAV is airborne, an actuator fault forces the UAV to perform an emergency landing, thus failing the mission. The probability of an actuator fault occurring in some period of time $T_a$ is denoted $P_a$. The probability of it occurring during any given time step is then defined by
\begin{align*}
	P_{a,dt} = 1 -(1 - P_a)^{\frac{dt}{T_a}}
\end{align*}

A fault in the grasping mechanism may also occur at any time. However, it only affects the mission when a target is tethered to the UAV. Upon this fault occurring, the target is dropped. The probability of a grasping mechanism fault occurring in some period of time $T_g$ is denoted $P_g$. The probability of it occurring during any given time step is then
\begin{align*}
	P_{g,dt}~=~1 - (1 - P_g)^{\frac{dt}{T_g}}
\end{align*}

Finally, we may define a generic \textit{system fault}. This represents the detection of some undesirable condition after the UAV has taken off. Examples of such conditions include interrupted communications between UAV and ground control station, erratic sensor feedback or the absence of some required system parameters or commands. The UAV's decision-making system has an \textit{Initialise} mode specifically for detecting these conditions. The probability that some undesirable condition is detected during this mode is given by $ P_s : \mathbb{R}_{\geq 0} \rightarrow [0,1] $ and is independent of time.

\section{Autonomous Guidance System}
\label{Sec:Guidance}
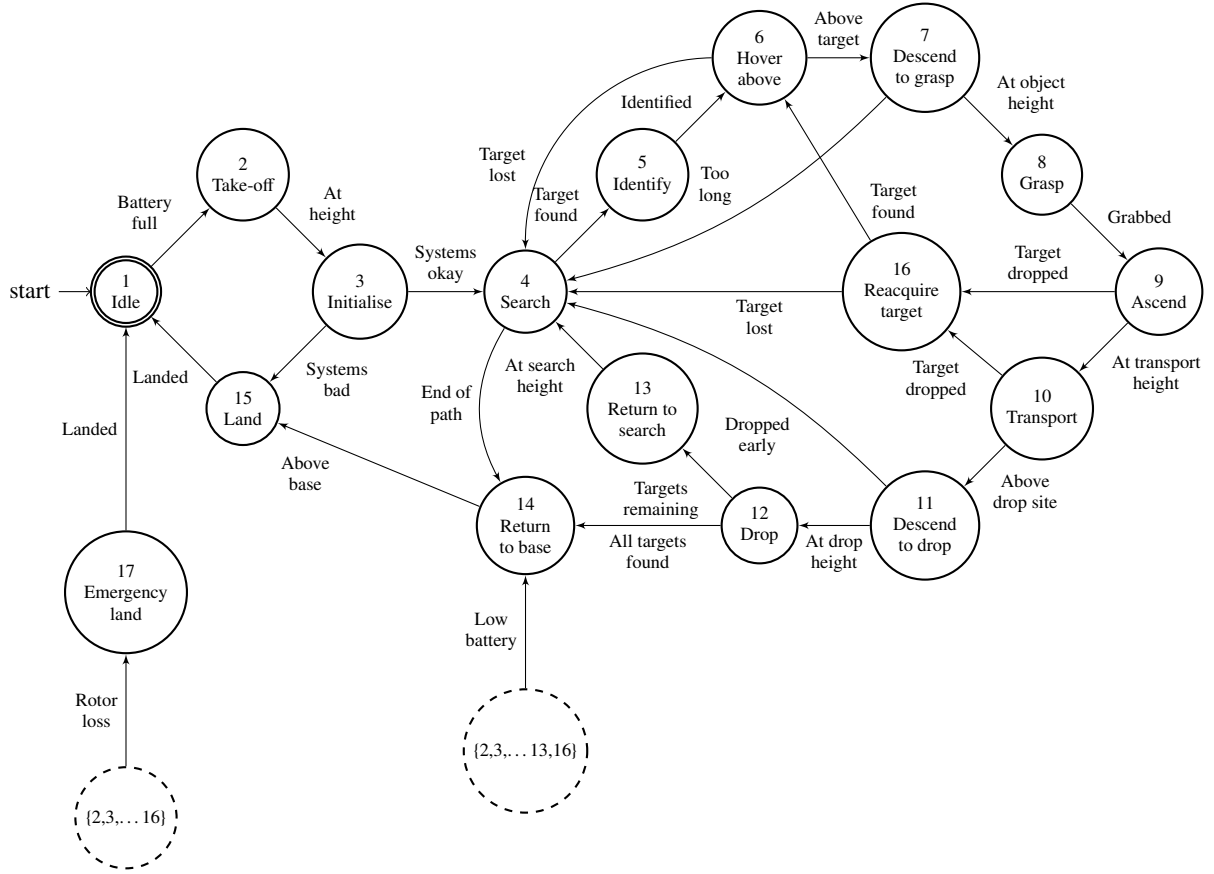
\begin{figure*}
	\centering
	\small
	\tikzstyle{line} = [draw, align=center, -latex', font=\scriptsize]
	\begin{tikzpicture}[node distance=2.2cm,on grid,auto,
	every state/.style={thick,font=\scriptsize}]


	\node[state,initial,accepting] (idle) [align=center] {1\\ Idle};
	\node[state] (takeoff) [above right of=idle, align=center] {2\\ Take-off};
	\node[state] (init) [below right of=takeoff, align=center] {3\\ Initialise};
	\node[state] (land) [below left of=init, align=center] {15\\ Land};
	\node[state] (search) [right of=init, align=center] {4\\ Search};
	\node[state] (identify) [above right of=search, align=center] {5\\ Identify};
	\node[state] (hover) [align=center, above right of=identify] {6\\ Hover\\ above};
	\node[state] (descendgrab) [align=center, right of=hover] {7\\ Descend\\ to grasp};
	\node[state] (grab) [below right of=descendgrab, align=center] {8\\ Grasp};
	\node[state] (ascend) [below right of=grab, align=center] {9\\ Ascend};
	\node[state] (transport) [below left of=ascend, align=center] {10\\ Transport};
	\node[state] (descenddrop) [align=center, below left of=transport] {11\\ Descend\\ to drop};
	\node[state] (drop) [left of=descenddrop, align=center] {12\\ Drop};
	\node[state] (rts) [align=center, below right of=search] {13\\ Return to\\ search};
	\node[state] (rtb) [align=center, below left of=rts] {14\\ Return\\ to base};
	\node[state] (reacquire) [align=center, right of=search, node distance=5cm] {16\\ Reacquire\\ target};
	\node[state] (emergency) [align=center, below of=idle, node distance=4cm] {17\\ Emergency\\ land};
	\node[state,dashed] (toemergency) [align=center, below of=emergency, node distance=3cm] {\{2,3,\ldots 16\}};
	\node[state,dashed] (tortb) [align=center, below of=rtb, node distance=3cm] {\{2,3,\ldots 13,16\}};

	\path[line]
	(idle) 			edge node {Battery\\ full} (takeoff)
	(takeoff) 		edge node {At\\ height} (init)
	(init) 			edge node {Systems\\ bad} (land)
	edge node {Systems\\ okay} (search)
	(search)		edge node [pos=0.6] {Target\\ found} (identify)
	edge [bend right] node [pos=0.5, left] {End of\\ path} (rtb)
	(identify)		edge node {Identified} (hover)
	(hover)			edge node {Above\\ target} (descendgrab)
	edge [bend right=45] node [pos=0.8, above left] {Target\\ lost} (search)
	(descendgrab)	edge node {At object\\ height} (grab)
	edge [bend left=15] node [pos=0.5, above left] {Too\\ long} (search)
	(grab)			edge node {Grabbed} (ascend)
	(ascend)		edge node {At transport\\ height} (transport)
	edge node [above] {Target\\ dropped} (reacquire)
	(transport)		edge node {Above\\ drop site} (descenddrop)
	edge node {Target\\ dropped} (reacquire)
	(descenddrop)	edge node {At drop\\ height} (drop)
	edge [bend right=15] node [below left, pos=0.3] {Dropped\\ early} (search)
	(drop)			edge node {Targets\\ remaining} (rts)
	edge node {All targets\\ found} (rtb)
	(rts)			edge node [pos=0.4] {At search\\ height} (search)
	(reacquire)		edge node [pos=0.1, above right] {Target\\ found} (hover)
	edge node [pos=0.3] {Target\\ lost} (search)
	(rtb)			edge node [pos=0.7] {Above\\ base} (land)
	(land) 			edge node [pos=0.3] {Landed} (idle)
	(emergency)		edge node {Landed} (idle)
	(toemergency)	edge node {Rotor\\ loss} (emergency)
	(tortb)			edge node {Low\\ battery} (rtb)
	;

	\end{tikzpicture}
	\caption{Finite-state machine describing AI logic.}
	\label{Fig:FSM}
\end{figure*}

The guidance system of the UAV consists of a number of autonomous \textit{modes} which dictate the behaviour of the UAV at any given time. Each mode comprises a combination of automatic control and trajectory modules, which collaborate to drive the UAV towards a specific position and heading. The modes and the transitions between them may be represented as a finite state machine, shown in Figure~\ref{Fig:FSM}. The UAV can respond to any internal or external event considered in the simulation. In a real-world scenario, events may occur which are unmodelled in the simulation. In implementing this guidance system in reality, this would make the UAV only semi-autonomous. Human intervention would likely be required to mitigate the effects of some events which the guidance system is not designed to react to.

Using the FSM to represent the behaviour of the guidance system, each mode corresponds to a state of the FSM. Only one mode may thus be active at any given time. The guidance system progresses from one mode to another under certain conditions. The criteria which dictate an exit from the current mode and which mode succeeds the current one are known as \textit{predicates}. The predicates for each mode are shown with reference to the FSM in Figure~\ref{Fig:FSM}.

We now detail the behaviour of each mode with reference to the control and trajectory modules or commands employed in each case. The algorithms of each of these modules are then described.

\subsection{Guidance Modes}
\label{Sec:Guidance_Modes}
The guidance modes determine which control and trajectory modules are active at any given time, and what the inputs to each of these modules are. Each mode thus comprises a unique set of instructions and modules. These are summarised in Table~\ref{Tab:Modules}.

\begin{table*}
	\newcommand{\nt}{\tabularnewline[0.1cm]}
	\newcommand{\nts}{\tabularnewline[-0.2cm]}
	\centering
	\caption{Guidance commands and modules for each AI mode.}
	\label{Tab:Modules}
	\begin{tabular}{l l l l}
		\toprule
		Mode	 & Input Command / Module & Trajectory Command / Module \\
		\midrule
		Idle	& $\vc{u}_\qud = [0,0,0,0]^T$ & -- \\
		Take-off	& State feedback controller & $\vc{r}_d = [x_0, y_0, z_\text{hvr}]^T$ \\
		Initialise & State feedback controller & $\vc{r}_d = [x_0, y_0, z_\text{hvr}]^T$ \\
		Search & State feedback controller & Search pattern \\
		Identify & State feedback controller & $\vc{r}_d = \vc{r}_\text{entry}$ \\
		Hover above target & State feedback, visual controllers & Object tracking, $z_d = z_\text{srch}$ \\
		Descend to grasp & State feedback, visual controllers & Object tracking, $z_d = -(z_{\grb/\qud} + 2R_\trg)$ \\
		Grasp & State feedback controller & $\dot{x}_d = 0, \dot{y}_d = 0, z_d = -(z_{\grb/\qud} + 2R_\trg)$  \\
		Ascend & State feedback controller & $\dot{x}_d = 0, \dot{y}_d = 0, z_d = z_\text{trnsprt}$  \\
		Transport & State feedback controller & $\vc{r}_d = [x_\ds,  y_\ds,z_\text{trnsprt}]^T$  \\
		Descend to drop & State feedback controller & $\vc{r}_d = [x_\ds,  y_\ds,  -(z_{\grb/\qud} + 2R_\trg)]^T$  \\
		Drop & State feedback controller & $\vc{r}_d = [ x_\ds,  y_\ds,  -(z_{\grb/\qud} + 2R_\trg)]^T$  \\
		Return to search & State feedback controller & $\vc{r}_d = [ x_\ds,  y_\ds, z_\text{srch}]^T$  \\
		Return to base & State feedback controller & $\vc{r}_d = [ x_0,  y_0, z_\text{hvr}]^T$  \\
		Reacquire target & State feedback controller & $\vc{r}_d = \vc{r}_\text{entry}$  \\
		Land & State feedback controller & $\vc{r}_d = [x_0,  y_0,-R_\qud]^T$ \\
		Emergency land & Emergency contoller & $z_d = -R_\qud$ \\
		\bottomrule
	\end{tabular}
\end{table*}

The UAV begins the mission in \textit{Idle}. It then progresses to \textit{Take-off} and subsequently \textit{Initialise}, where a self-diagnosis is performed. The probability of continuing the mission (as opposed to landing) is determined by the system fault probability $P_s$. The UAV may then either progress to \textit{Search} or \textit{Land} mode, whereupon it continues as per the predicates of each state. The modes, transitions and predicates are detailed in Figure~\ref{Fig:FSM}. Each mode may be summarised as follows.

\subsubsection{Idle}
The UAV sits at rest with its rotors inactive. The UAV begins the mission in this mode, resting at the landing site. It returns to this mode after completing the mission, with one of two landing modes leading to \textit{Idle}. When succeeding \textit{Land}, the UAV rests at the landing site. When succeeding \textit{Emergency land}, its location is wherever it comes to rest after colliding with the floor.

The UAV exists this mode only when $V = V_\text{max}$. This ensures that the battery is fully charged before beginning or resuming the mission. It may then progress to \textit{Take-off} if the mission has not yet been completed.

\subsubsection{Take-off}
The UAV takes off to a position above the landing site $\vc{r}_d = [x_0,y_0,z_\text{hvr}]^T$, where $z_\text{hvr}$ is the hover height. This ensures that the ground effect region is cleared before performing further manoeuvres. It then proceeds to \textit{Initialise} upon satisfying the condition $\| \vc{r}_\qud - \vc{r}_d \| < tol $ where $tol$ is a tolerance value which varies from case to case.

\subsubsection{Initialise}
The UAV performs a self-diagnosis to identify any system faults. A system fault is detected with probability $P_s$, in which case the UAV proceeds to \textit{Land}. If no fault is detected, the UAV makes a transition to \textit{Search}.

\subsubsection{Search}
The UAV follows a series of waypoints as defined by the \textit{Search pattern} trajectory module. The spacing of the waypoints and the search height $z_\text{srch}$ are defined such that the camera's field of view overlaps as it passes between pairs of waypoints. The velocity command is limited to $v_{\max} = 2$ \si{\metre/\second} to ensure that the UAV does not pass over targets too quickly. The \textit{Object tracking} module is active during this mode, but does not inform the controller. It provides only a boolean $D \in \{0,1\}$ which specifies whether a target has been detected. When $D=1$, a transition to \textit{Identify} is triggered (see Figure~\ref{Fig:FSM}).

As the UAV is aware of how many targets are present in the environment, it will return to base if all targets have been located and retrieved. In such a situation, it does not therefore reach the final waypoint. If it reaches the final waypoint, it is assumed that not all targets have been found. The UAV then proceeds to \textit{Return to base} and triggers a \textit{Mission failed} flag. A time limit $T_\text{max}$ defined for the mission triggers a similar behaviour.

\subsubsection{Identify}
The transition to \textit{Identify} is triggered when a target is detected by the \textit{Object tracking} module, as described in Section~\ref{Sec:Object_Tracking}. The UAV typically has some lateral motion during this transition. The controller thus compensates for overshooting the target by returning the UAV to the position recorded as it entered \textit{Identify}, denoted $\vc{r}_\text{entry}$. Once it has returned within a small radius of this position, that is $\|\vc{r}_\qud - \vc{r}_\text{entry}\| < tol$, the UAV progresses to \textit{Hover above target}.

\subsubsection{Hover Above Target}
\begin{figure}
	\centering
	\includegraphics[scale=0.7]{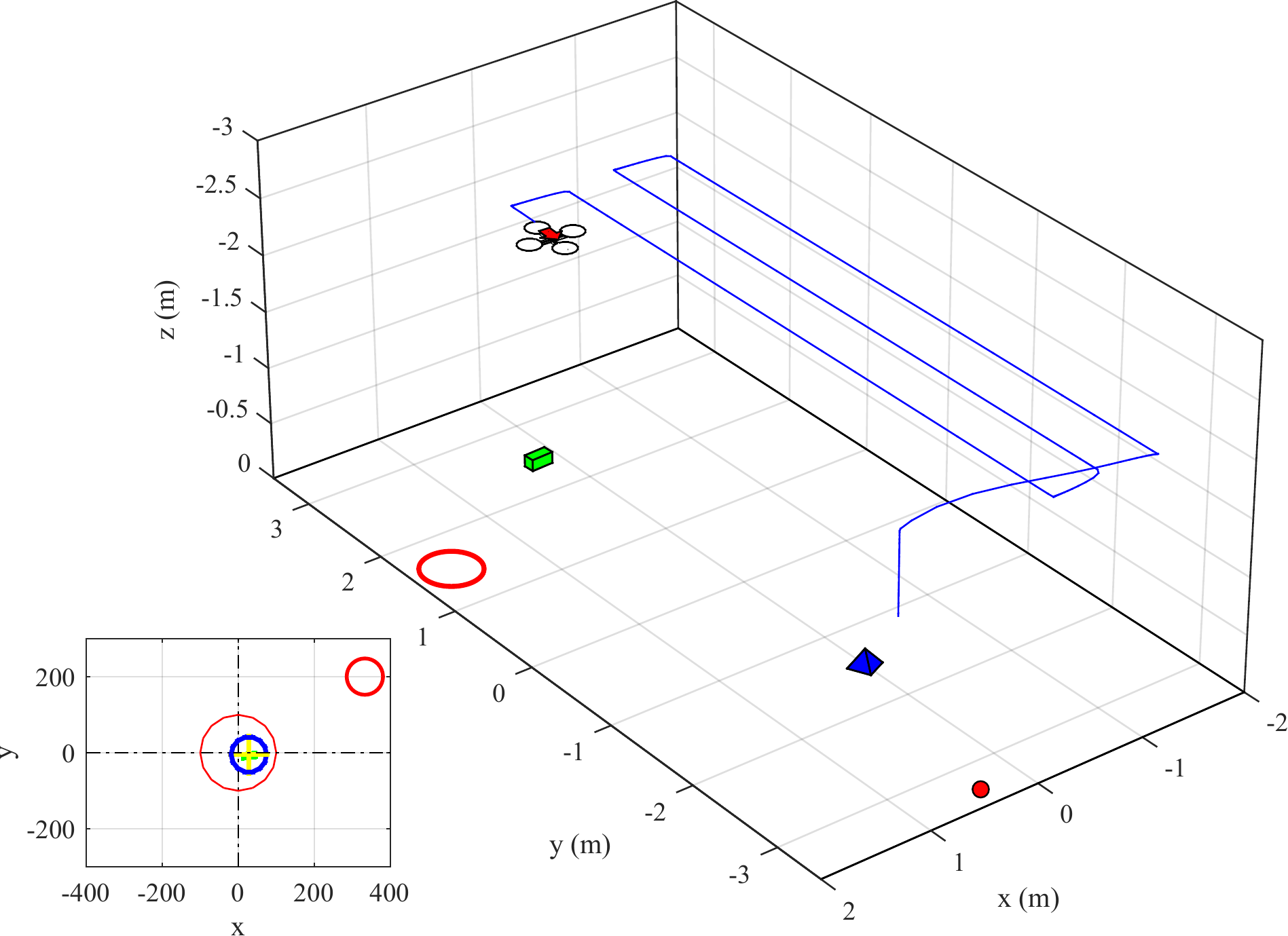}
	\caption{Mission visualisation, showing UAV, targets, UAV flight path and camera sensor image.}
	\label{Fig:Camera_View}
\end{figure}

As a result of the position command issued during \textit{Identify}, the UAV is now in a position above the target which makes the target visible to the camera sensor (see Figure~\ref{Fig:Camera_View}). The \textit{Object tracking} module detects blocks of colour within a certain RGB range and determines the coordinates of their centroids. The coordinates of the centroid closest to the centre of the image are supplied to the \textit{Visual controller}. This has the result of moving the UAV directly above the target. When the vehicle velocity in the $x^\Wframe$-$y^\Wframe$ plane is near-zero, that is $\sqrt{\dot{x}^2_\qud + \dot{y}^2_\qud} < tol$, the UAV is above the target and the UAV proceeds to \textit{Descend to grasp}.

\subsubsection{Descend to Grasp}
The UAV uses the \textit{Object tracking} and \textit{Visual controller} modules to maintain its position directly above the target, while descending to retrieve it. When the grasping mechanism position is within a small radius of the top of the target, that is $\|\vc{r}_\grb - (\vc{r}_\trg - R_\trg)\| < tol$, the UAV makes a transition to \textit{Grasp}.

\subsubsection{Grasp}
The grasping mechanism is activated and the dynamics of the grasped target are tethered to those of the UAV, as described by Equation \eqref{Eq:Target_Dynamics_Tethered}. The UAV then proceeds to \textit{Ascend}.

\subsubsection{Ascend}
The UAV ascends to the transport height $z_\text{trnsprt}$, while maintaining zero velocity in the horizontal plane. The UAV proceeds to \textit{Transport} upon reaching $z_\text{trnsprt}$, that is, satisfying $|z_\qud - z_\text{trnsprt}| < tol$. A grasping mechanism fault occurs in this mode with probability $P_g$. When such a fault is detected, the UAV moves instead to \textit{Reacquire target}.

\subsubsection{Transport}
The UAV transports the tethered target to a position above the drop site, defined by $\vc{r}_d = [x_\ds,y_\ds,z_\text{trnsprt}]^T$. The UAV proceeds to \textit{Descend to drop} upon the UAV reaching this position, that is $\|\vc{r}_\qud - \vc{r}_d\| < tol$, for the mode-specific command $\vc{r}_d$. A grasping mechanism fault occurs in this mode with probability $P_g$. Again, when such a fault is detected, a transition to \textit{Reacquire target} is triggered.

\subsubsection{Descend to Drop}
The UAV descends to a height where the target is touching the floor, that is $z_d = -(z_{\grb/\qud} + 2R_\trg)$. If the target is dropped early during this mode due to a grasping mechanism fault, the target is recorded as having been successfully deposited and the UAV proceeds to \textit{Return to search}. Otherwise, it progresses to \textit{Drop} upon satisfying $\|\vc{r}_\qud - \vc{r}_d\| < tol$, where $\vc{r}_d = [x_\ds,y_\ds,z_d]^T$.

\subsubsection{Drop}
The grasping mechanism is deactivated and the dynamics of the target are untethered from the UAV and described again by Equation \eqref{Eq:Target_Dynamics_Untethered}. The UAV proceeds to \textit{Return to search} if the mission is not complete (i.e.\ there are targets still to be found) or to \textit{Return to base} if all targets have been deposited at the drop site.

\subsubsection{Return to Search}
The UAV ascends vertically to search height above the drop site, that is, the position $\vc{r}_d = [x_\ds,y_\ds,z_\text{srch}]^T$. It then proceeds to \textit{Search} upon satisfying $\|\vc{r}_\qud - \vc{r}_d\| < tol$.

\subsubsection{Return to Base}
The UAV returns to a position $\vc{r}_d = [x_0,y_0,z_\text{hvr}]^T$ above the landing site upon either successfully depositing all targets, reaching the end of the search path or receiving a low battery warning. A transition to \textit{Land} is triggered upon satisfying $\|\vc{r}_\qud - \vc{r}_d\| < tol$.

\subsubsection{Land}
The UAV descends from its position above the landing site to rest on the landing site itself, at position $\vc{r}_d = [x_0,y_0,-R_\qud]^T$. The UAV then returns to \textit{Idle} upon satisfying the conditions $\|\vc{r}_\qud - \vc{r}_d\| < tol_1$ and $\|\dot{\vc{r}}_\qud\| < tol_2$. This ensures that the motors are set to idle when the UAV is stationary at the landing site.

\subsubsection{Reacquire Target}
A transition to \textit{Reacquire target} from \textit{Ascend} or \textit{Transport} is triggered upon a grasping mechanism fault occurring. The transition occurs immediately upon the grasping mechanism failing. As the UAV may have a high velocity when the transition occurs, it is commanded to return to the position $\vc{r}_\text{entry}$ at which it entered the \textit{Reacquire target} mode in order to maximise the probability of visually reacquiring the target. To exit this mode, the conditions $\| \vc{r}_\qud - \vc{r}_\text{entry}\| < tol_1$, $\| \dot{\vc{r}}_\qud \| < tol_2$ must be satisfied. The guidance mode changes to \textit{Hover above target} if the \textit{Object tracking} module detects a target, that is, the module has set $D=1$, or to \textit{Search} if the previous conditions are satisfied and no target is detected, that is, $D=0$. That is, the target has been lost and the highest probability of reacquiring it occurs while following the search waypoints.

\subsubsection{Emergency Land}
A transition to \textit{Emergency land} may occur in any mode bar \textit{Emergency land} itself and \textit{Idle}. The \textit{Emergency controller} is employed, accepting only the height command $z_d = -R_\qud$. The UAV returns to \textit{Idle} upon the condition $| z_\qud + R_\qud | < tol$ being satisfied.

\subsection{Additional Transitions}
As highlighted by Figure~\ref{Fig:FSM}, the modes \textit{Reacquire target} and \textit{Emergency land} may be reached from a large number of other modes. These transitions are triggered by the detection of faults. The autonomous controller reacts to the occurrence of faults by switching to a mode which is designed to handle the effects of the fault.

\subsubsection{Low Battery Warning}
A simple battery monitor subsystem checks the battery voltage $V$ and triggers a transition to \textit{Return to base} when $V \leq V_\text{th}$, where $V_\text{th}$ is the lower threshold.  This transition can occur in any mode except those in the path leading from \textit{Return to base} to \textit{Idle}, that is \textit{Return to base}, \textit{Land}, \textit{Emergency land} or \textit{Idle}. Following this transition, any tethered targets are released, as the mission priority is to recharge the battery.

\subsubsection{Actuator Fault}
An actuator fault occurs with probability $P_a$. This triggers a transition to \textit{Emergency land}. It may occur in any mode except \textit{Emergency land} or \textit{Idle}.

\subsection{Control and Trajectory Modules}
\label{Sec:Control_Modules}
The control and trajectory modules active at any given time are determined by the UAV's mode, as previously described. The modules work in combination to follow the commands issued in the active mode. As an example, during \textit{Take-off}, the desired vehicle behaviour is to hover at a static position above the landing site. A constant position command is therefore supplied to the \textit{State feedback controller}. In \textit{Search}, the UAV is required to follow one waypoint after another. It therefore requires a \textit{Search pattern} module to define the changing position commands and the \textit{State feedback controller} to track them. A \textit{State reconstruction} module is active at all times. This module is required to reconstruct the dynamic states of the quadrotor from the available sensor measurements.

The vehicle control laws determine the \textit{pseudo-inputs} of the UAV, $\hat{\vc{u}} = [u_\col,u_\lat,u_\lon,u_\yaw]^T$, detailed in \cite{Ireland2015,Ireland2015a,Voos2009}. We may define these four pseudo-inputs in terms of the true rotor inputs $\vc{u} = [u_1,u_2,u_3,u_4]^T$ such that
\begin{equation}
	\hat{\vc{u}} = \vc{C} \vc{u}, \quad
	\text{where } \vc{C} = \begin{bmatrix*}[r]
		1 & 1 & 1 & 1 \\
		0 & 0 & 1 & -1 \\
		-1 & 1 & 0 & 0 \\
		-1 & -1 & 0 & 0
	\end{bmatrix*}
\end{equation}
If the controller determines the pseudo-inputs, then the true rotor inputs are found from the inverse relationship $\vc{u}~=~\vc{C}^{-1} \hat{\vc{u}}$.

This section describes the control and trajectory modules which work in combination to supply these pseudo-inputs and therefore drive the UAV behaviour. The module combinations and mode-dependent commands are specified in Table~\ref{Tab:Modules}.

\subsubsection{State Reconstruction}
The control and navigation systems of the UAV require knowledge of the system states $\vc{x}_\qud$. These are not directly obtainable and must be measured through the sensors. The states may then be reconstructed in software from the sensor outputs. The sensor model omits errors in the measurement relationships. It is therefore assumed that the position $\vc{r}_\qud$ and attitude $\vc{\eta}_\qud$ are perfectly reconstructed from the Optitrack measurements, while the angular rates $\vc{\omega}_\qud$ are similarly reconstructed without error from the gyroscope outputs. Translational velocity is obtained by filtering and differentiating the vehicle position
\begin{equation}
	\hat{\vc{v}}_\qud = \frac{Ns}{s+N} \vc{r}_\qud
\end{equation}
where $N$ is the filter bandwidth.

\subsubsection{State Feedback Controller}
A state feedback controller with feedback linearisation \cite{Ireland2015,Voos2009,Das2009} stabilises the quadrotor and ensures accurate tracking of position and yaw commands. The position controller consists of cascaded state feedbacks, with the desired acceleration given by
\begin{equation}
	\ddot{\vc{r}}_d = K_{dr} \left( \dot{\vc{r}}_d - \dot{\vc{r}} \right)
	\label{Eq:Velocity_Controller}
\end{equation}
and the desired velocity given by
\begin{equation}
	\dot{\vc{r}}_d = K_{pr} \left( \vc{r}_d - \vc{r}_\qud  \right)
	\label{Eq:Position_Controller}
\end{equation}
where $\vc{r}_d$ is the desired position and $\dot{\vc{r}}_d$ is subject to the limit $\|\dot{\vc{r}}_d \| \leq v_\text{max}$, where $v_\text{max}$ is an artificial maximum velocity limit. Decoupling of the position and velocity feedbacks is required for the visual controller described by Equation \eqref{Eq:Visual_Controller}.
Linearising feedbacks then use the desired acceleration command $\ddot{\vc{r}}_d$ to determine the collective pseudo-input
\begin{equation}
	u_\col = \frac{m_\qud(g - \ddot{z}_d)}{K_T\cos\phi\cos\theta}
\label{Eq:Collective_Input}
\end{equation}
and the roll and pitch commands
\begin{equation}
\begin{aligned}
	\phi_d &= \arcsin \left( \frac{m_\qud\left( \ddot{y}_d \cos\psi - \ddot{x}_d \sin\psi \right)}{K_T u_\col} \right) \\
	\theta_d &= -\arcsin \left( \frac{m_\qud\left( \ddot{x}_d \cos\psi + \ddot{y}_d \sin\psi \right)}{K_T u_\col \cos\phi} \right)
\end{aligned}
\end{equation}
where the commands are subject to the limits $\{|\phi_d|, |\theta_d|\} \leq a_{\max}$, where $a_\text{max}$ is an artificial maximum roll/pitch limit. The attitude controller is similarly defined by the state feedback
\begin{equation}
	 \dot{\vc{\omega}}_d = \vc{K}_{p\eta} \left( \vc{\eta}_d - \vc{\eta}_\qud \right) - \vc{K}_{d\eta} \vc{\omega}_\qud
\end{equation}
and the linearising feedbacks
\begin{equation}
	\begin{bmatrix}
		u_\lat \\ u_\lon \\ u_\yaw
\end{bmatrix} = \begin{bmatrix}
	\frac{I_x}{K_TL} & 0 & 0 \\
	0 & \frac{I_y}{K_TL} & 0 \\
	0 & 0 & \frac{I_z}{K_Q}
\end{bmatrix} \dot{\vc{\omega}}_d
\end{equation}
where $\vc{K}_{p\eta} = [K_{p\phi}, K_{p\theta}, K_{p\psi}]$ and $\vc{K}_{d\eta} = [K_{d\phi}, K_{d\theta}, K_{d\psi}]$.

\subsubsection{Search Pattern}
The search pattern module defines a series of $n_w$ waypoints which the UAV follows after taking off. The waypoint locations are defined such that following each one from the beginning to the end of the search pattern results in near-exhaustive visual coverage of the environment floor. As each waypoint is reached, the counter $c_\text{wp} \in \{1,2,\ldots, n_w\}$ is incremented. Then, if the UAV exits \textit{Search} to investigate a target or return to base, it is able to return to the last waypoint reached. This ensures comprehensive coverage of the environment. The waypoints are listed in Table~\ref{Tab:Waypoints}.

\begin{table}
	\centering
	\caption{Waypoints for UAV search mode.}
	\label{Tab:Waypoints}
	\begin{tabular}{c r r r r}
		\toprule
		Waypoint & $x_d$ & $y_d$ & $z_d$ & $\psi_d$ \\
		\midrule
		1 & $-1.5$ & $-3.0$ & $-2$ & $\frac{\pi}{2}$ \\
		2 & $-1.5$ & $3.0$ & $-2$ & $\frac{\pi}{2}$ \\
		3 & $-1.0$ & $3.0$ & $-2$ & $-\frac{\pi}{2}$ \\
		4 & $-1.0$ & $-3.0$ & $-2$ & $-\frac{\pi}{2}$ \\
		5 & $-0.5$ & $-3.0$ & $-2$ & $\frac{\pi}{2}$ \\
		6 & $-0.5$ & $3.0$ & $-2$ & $\frac{\pi}{2}$ \\
		7 & $0.0$ & $3.0$ & $-2$ & $-\frac{\pi}{2}$ \\
		8 & $0.0$ & $-3.0$ & $-2$ & $-\frac{\pi}{2}$ \\
		9 & $0.5$ & $-3.0$ & $-2$ & $\frac{\pi}{2}$ \\
		10 & $0.5$ & $3.0$ & $-2$ & $\frac{\pi}{2}$ \\
		11 & $1.0$ & $3.0$ & $-2$ & $-\frac{\pi}{2}$ \\
		12 & $1.0$ & $-3.0$ & $-2$ & $-\frac{\pi}{2}$ \\
		13 & $1.5$ & $-3.0$ & $-2$ & $\frac{\pi}{2}$ \\
		14 & $1.5$ & $3.0$ & $-2$ & $\frac{\pi}{2}$ \\
		\bottomrule
	\end{tabular}
\end{table}

\subsubsection{Emergency Controller}
The emergency controller is similar to the state feedback controller in structure. It is utilised only when an actuator fault occurs and an emergency landing is required. In this event, it is assumed that a single rotor has malfunctioned and it is known which rotor this is. The controller then disables the opposing rotor and increases the thrust to the remaining two rotors to reduce the chance of a hard landing.

With one rotor inoperable, attitude control is neglected to avoid increasing the possibility of a crash. The collective pseudo-input relationship is still given by Equation \eqref{Eq:Collective_Input}. The remaining pseudo-inputs are fixed at zero. With two rotors inactive, the remaining two rotors must each bear half of the required thrust, giving
\begin{equation}
	\vc{u}_\qud = \begin{cases}
		[0.5, 0.5, 0, 0]^T u_\col & \text{if rotor 3 or 4 is faulty} \\
		[0, 0, 0.5, 0.5]^T u_\col & \text{if rotor 1 or 2 is faulty}
	\end{cases}
\end{equation}

\subsubsection{Object Tracking}
\label{Sec:Object_Tracking}
The object tracking module is used to identify targets within the camera's field of view. The centroid of an identified object in the camera image provides coordinates which are utilised by the visual controller. Thus, upon identifying a target, the UAV is able to accurately position itself above the target and descend to collect it.

The object recognition subsystem emulates that of a simple colour detection algorithm. The targets have distinctive colours which allow them to stand out from each other and the environment. The subsystem masks certain colour ranges in the camera image, thus allowing detection of the targets when they enter the camera field of view. This is emulated in simulation by considering which vertices of the global geometry are within the camera field of view, then identifying the faces and colours associated with these vertices. The actual process of colour detection and the simulated process are described in this section.

\paragraph{Colour Detection Approach}
A colour image of resolution $p\times q$ is typically described by a $p \times q \times 3$ array. Different colour models may be used. In this instance, the popular RGB (red-green-blue) model is employed. Each of the three $p \times q$ slices of the array then describes the intensity of the image in red, green and blue. Intensity values typically vary between 0 and 255. Colour detection is achieved by considering ranges of values for each of the three colours.

\begin{figure}
	\newcommand{\figwidth}{5cm}
	\centering
	\begin{subfigure}[t]{0.32\columnwidth}
		\centering
		\includegraphics[width=\textwidth]{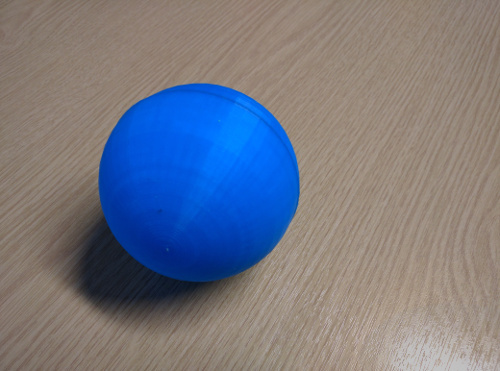}
		\caption{Captured image.}
		\label{Fig:OD_Image}
	\end{subfigure}\;
	\begin{subfigure}[t]{0.32\columnwidth}
		\centering
		\includegraphics[width=\textwidth]{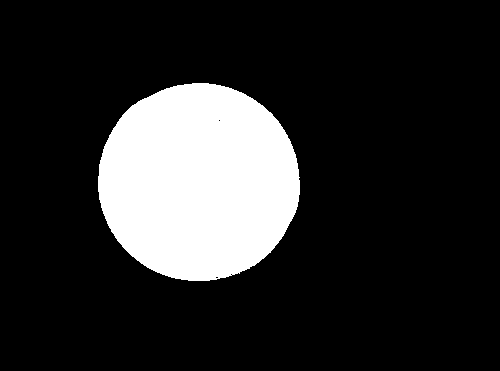}
		\caption{Image mask.}
		\label{Fig:OD_Mask}
	\end{subfigure}\;
	\begin{subfigure}[t]{0.32\columnwidth}
		\centering
		\includegraphics[width=\textwidth]{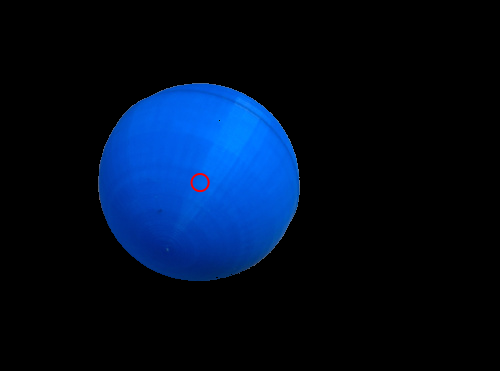}
		\caption{Mask applied to image, with centroid indicated.}
		\label{Fig:OD_MaskedImage}
	\end{subfigure}
	\caption{Mask applied to image and centroid indicated.}
\end{figure}

Consider the image of a simple monochromatic object, shown in Figure~\ref{Fig:OD_Image}. A mask is generated, which shows only the colours within a certain range, as shown in Figure~\ref{Fig:OD_Mask}. The centroid of the unmasked region then provides the approximate geometrical centre of the object, as shown in Figure~\ref{Fig:OD_MaskedImage}.

The target objects are defined using distinctive colours which are unique in the environment. The coordinates of the centroid then provide the feedback for the visual controller. This allows the UAV to position itself according to the imagery received by the camera, as interpreted by the colour detection algorithm.

\paragraph{Object Detection Model}
The geometry of the model is described in Section~\ref{Sec:Geometry} and the transformation of vertices into camera space described in Section~\ref{Sec:Camera_Sensor}. To be detected, at least one vertex of a single object must satisfy the conditions for visibility to the camera sensor.

Consider the case where $m_\text{vis}$ vertices are visible to the camera, with coordinates $\vc{c}_i, i \in \mathbb{N}^{m_\text{vis}} \subseteq \{1,2,\ldots,m\}$, $m$ being the total number of vertices in the geometrical model of the scenario. That is, each set of coordinates $\vc{c}_i$ satisfies the conditions imposed by Equation \eqref{Eq:Camera_Limits}. Each set of coordinates corresponds to a vertex $\vc{v}_i$. Each vertex is then associated with one or more faces $\vc{f}_j \subseteq \{1,2,\ldots,m\}$. Thus, each vertex is also associated with one more colours $\vc{C}_j \in [0,1]$, where the RGB range $[0,255]$ is normalised in the range $[0,1]$.

Colour detection is achieved by considering only vertices with an associated colour inside a range. For example, to detect red objects, the colour arrays $\vc{C}_j$ associated with all visible coordinates $\vc{c}_i$ are considered in the limits $\vc{C}_\text{lw} = [0.4,0,0]^T$, $\vc{C}_\text{up} = [1,0,0]^T$. In the simplified geometry we employ in our model, this ignores all colours except the solid red which represents the spherical target. We then consider the $b$ coordinates $\vc{c}_i, i \in \mathbb{R}^b$ for which the associated colours satisfy the condition $\vc{C}_\text{lw} \leq \vc{C}_j \leq \vc{C}_\text{up}$.

Hence, any colour of interest is detected and its vertices identified. We then use a simple centroid calculation to obtain the approximate centre of the target in the camera image
\begin{equation}
	\vc{c}_\text{cntrd} = \frac{1}{b}\sum_{i=1}^b \vc{c}_i
\end{equation}

The detection of a target is specified by the boolean $D \in \{0,1\}$.
To avoid detecting targets at the extremes of the image and then losing them due to the UAV motion, a target is only considered as having been identified when its centroid is within a certain radius of the image centre. This boundary is clearly visible in Figure~\ref{Fig:Camera_View}. Thus $D = 1$ if $\| \vc{c}_\text{cntrd} \| < R_c$, where $R_c$ is the radius around the image centre.

Targets already in the drop zone are also ignored. This is accomplished by identifying the drop site position in the camera image $ \vc{c}_\ds $ and rejecting centroids outside the radius $ R_{\ds,\cam} $. Hence $ D = 0 $ if $ \| \vc{c}_\ds \| < R_{\ds,\cam} $, where $ R_{\ds,\cam} $ approximates the drop zone radius by
\begin{align*}
	R_{\ds,\cam} = f\frac{R_\ds}{| z_\cam |}
\end{align*}

The object detection algorithm checks for all three primary colours simultaneously. Where multiple centroids are defined, that is, multiple colours are visible to the camera, the centroid closest to the image centre is supplied to the visual controller. This centroid $\vc{c}_\text{cntrd}$ is supplied to the visual controller.

\subsubsection{Visual Controller}
The coordinates of the centroid are denoted $\vc{c}_\text{cntrd} = [c_x,c_y]$. The position controller acts to drive the coordinates towards the centre of the camera image. To achieve consistency with the state feedback controller, the error is defined as a function of the coordinates and the camera height above the ground, that is
\begin{equation}
	\vc{e} = \left| \frac{z_\qud - z_{\cam/\qud}}{f} \right| \begin{bmatrix}
		c_y \\ c_x
	\end{bmatrix}
\end{equation}
where $z_{\cam/\qud}$ is the component of the camera position in $\Qframe$ and the centroid coordinates are swapped for consistency with the reference frame $\Qframe$.

A proportional-integral (PI) controller then centres the coordinates in the camera image and the target directly below the UAV. The controller provides a desired velocity command which is simply supplied to the state feedback controller in lieu of the existing horizontal position controller and is given by
\begin{equation}
	\begin{bmatrix}
		\dot{x}_d \\ \dot{y}_d
	\end{bmatrix} = \begin{bmatrix}
		\cos\psi & -\sin\psi \\
		\sin\psi & \cos\psi
	\end{bmatrix} \left( K_{pv} \vc{e} + K_{iv} \int \vc{e}\, dt \right)
\label{Eq:Visual_Controller}
\end{equation}
where $K_{pv}$, $K_{iv}$ are the proportional and integral gains, respectively.

These commands override the horizontal velocity commands of the state feedback controller in Equation \eqref{Eq:Velocity_Controller}. The height of the UAV continues to be controlled by the position controller (Equation \eqref{Eq:Position_Controller}) in this mode.

\section{Results}
\label{Sec:Results}
The model is used in a Monte Carlo experiment. The stochastic nature of the model has the effect of changing the timing and outcome of events and the ultimate success or failure of the mission. In this section, we first present and discuss the results of a single simulation run. The results of a Monte Carlo simulation conducted over 2,000 runs are then considered. The agent models employ the properties described in Table~\ref{Tab:Sim_Properties}. The UAV's guidance has the parameters described in Table~\ref{Tab:Guidance_Properties}.

\begin{table*}
	\centering
	\caption{Quadrotor, target and environment properties.}
	\label{Tab:Sim_Properties}
	\begin{tabular}{l c r l}
		\toprule
		Property										& Symbol	& Value	& Unit \\
		\midrule
		Camera aspect ratio								& $A$ & 4/3 & -- \\
		Quadrotor/floor damping coefficient 			& $c_\qud$	& 75.5 & \si{\newton \second\per\metre} \\
		Target/floor damping coefficient 				& $c_\trg$	& 20 & \si{\newton \second\per\metre} \\
		Camera focal length 							& $f$	& 400	& -- \\
		Acceleration due to gravity						& $g$	& 9.81	& \si{\metre\per\second\squared} \\
		Quadrotor moment of inertia about $x^\Qframe$ 	& $I_x$ & 0.03 & \si{\kilogram \square\metre} \\
		Quadrotor moment of inertia about $y^\Qframe$ 	& $I_y$ & 0.03 & \si{\kilogram \square\metre} \\
		Quadrotor moment of inertia about $z^\Qframe$ 	& $I_z$ & 0.04 & \si{\kilogram \square\metre} \\
		Rotor torque gain								& $K_Q$ & 120 & \si{\newton} \\
		Rotor thrust gain								& $K_T$ & 0.2 &	\si{\newton \metre} \\
		Quadrotor/floor spring coefficient 				& $k_\qud$ & 3775 & \si{\newton\per\metre} \\
		Target/floor spring coefficient 				& $k_\trg$ & 1000 & \si{\newton\per\metre} \\
		Rotor moment arm								& $L$	& 0.2 & \si{\metre} \\
		Quadrotor mass 									& $m_\qud$ & 1.51 & \si{\kilogram} \\
		Target mass 									& $m_\trg$ & 0.4 & \si{\kilogram} \\
		Probability of actuator fault in period $T_a$ 	& $P_a$ & 0.01 & -- \\
		Probability of grasper fault in period $T_g$ 	& $P_g$ & 0.05 & -- \\
		Probability of system fault during \textit{Initialise} & $P_s$ & 0.05 & -- \\
		Drop zone radius								& $R_\ds$	& 0.25	& \si{\metre} \\
		Quadrotor effective radius						& $R_\qud$	& 0.2 & \si{\metre} \\
		Target effective radius							& $R_\trg$	& 0.05 & \si{\metre} \\
		Camera position in $\Qframe$					& $\vc{r}_{\cam/\qud}$	& $[0,0,0.1]^T$ & \si{\metre} \\
		Grasper position in $\Qframe$					& $\vc{r}_{\grb/\qud}$	& $[0, 0, 0.2]^T$ & \si{\metre} \\
		Time period for actuator fault probability $P_a$ & $T_a$ & 60 & \si{\second} \\
		Time period for actuator fault probability $P_g$ & $T_g$ & 60 & \si{\second} \\
		Maximum battery voltage 						& $V_\text{max}$ & 11 & \si{\volt} \\
		Battery charging rate 							& $v_c$ & 0.025 & \si{\volt\per\second} \\
		Battery discharging rate 						& $v_d$ & -0.005 & \si{\volt\per\second} \\
		Time constant of gimbal response				& $\tau_g$	& 0.005 & \si{\second} \\
		Camera field of view							& $\lambda$	& 45	& \si{\degree} \\
		\bottomrule
	\end{tabular}
\end{table*}

\begin{table}
	\centering
	\caption{Guidance system and controller properties.}
	\label{Tab:Guidance_Properties}
	\begin{tabular}{l c r l}
		\toprule
		Property	& Symbol	& Value	& Unit \\
		\midrule
		Maximum roll/pitch command 	& $a_\text{max}$ & 20 & \si{\degree} \\
		Velocity controller gain	& $K_{dr}$	& 3.9	& -- \\
		Roll/pitch rate controller gain	& $K_{d\phi},K_{d\theta}$ & 39 & -- \\
		Yaw rate controller gain	& $K_{d\psi}$	& 1.95 & -- \\
		Visual controller integral gain	& $K_{iv}$ & 0.0001 & -- \\
		Position controller gain 	& $K_{pr}$ 	& 0.975 	& -- \\
		Visual controller proportional gain	& $K_{pv}$	& 0.341 & -- \\
		Roll/pitch angle controller gain	& $K_{p\phi},K_{p\theta}$ & 380.25 & -- \\
		Yaw angle controller gain	& $K_{p\psi}$	& 0.951 & -- \\
		Velocity filter bandwidth	& $N$	& 50 & \si{\radian\per\second} \\
		Maximum velocity command & $v_\text{max}$ & 5 & \si{\metre\per\second} \\
		Hover height & $z_\text{hvr}$ & -1 & \si{\metre} \\
		Search height & $z_\text{srch}$ & -2 & \si{\metre} \\
		Transport height & $z_\text{trnsprt}$ & -1 & \si{\metre} \\
		\bottomrule
	\end{tabular}
\end{table}

\subsection{Single Run}
The simulation is run with landing site, drop site and target positions defined at random and the potential for faults determined by the corresponding probability distribution. In this instance, the mission is successful and takes 170 \si{\second} to complete.

\begin{figure*}
	\centering
	\includegraphics[scale=0.8]{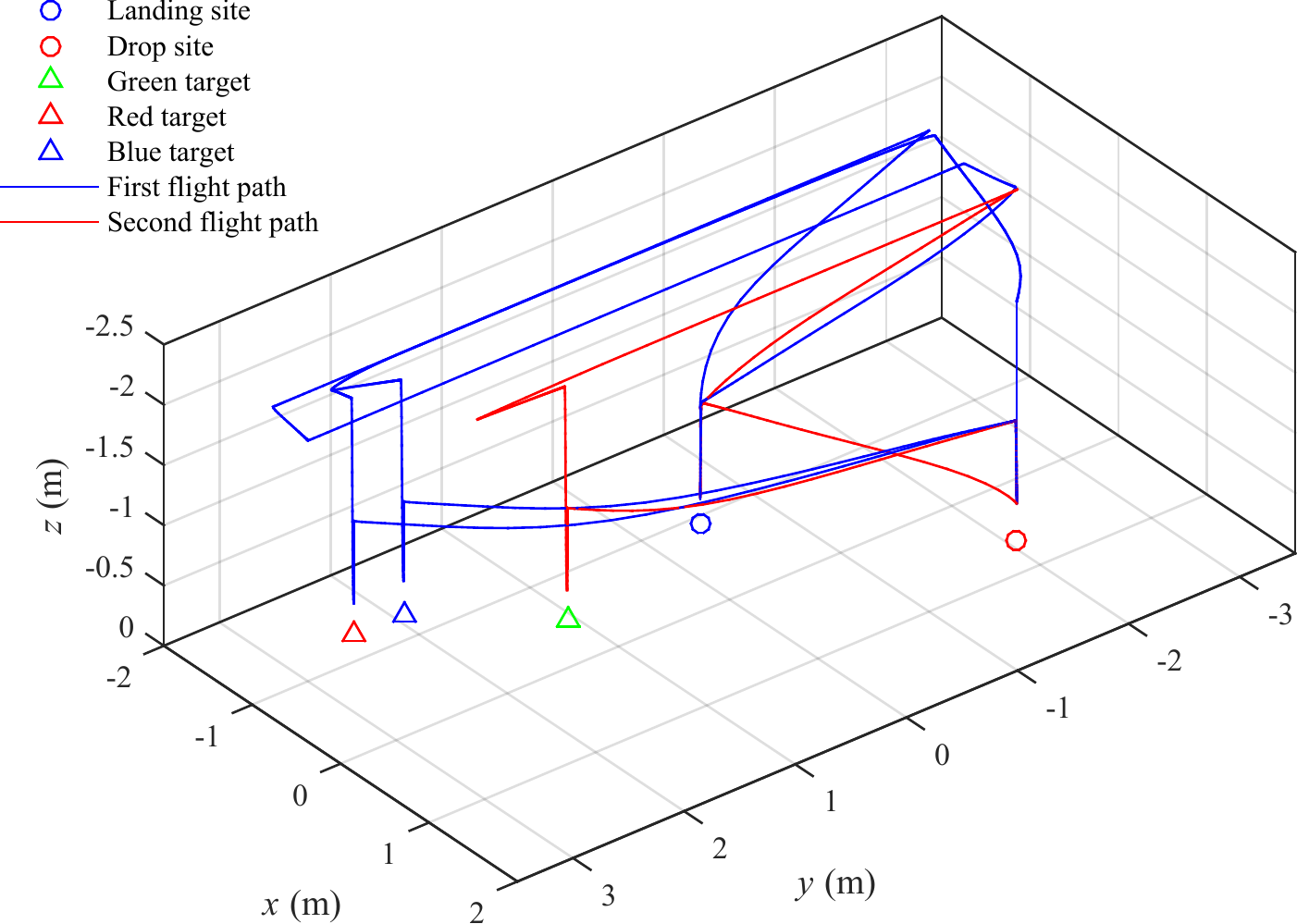}
	\caption{UAV flight path and target position during example mission.}
	\label{Fig:Single_Above}
\end{figure*}

We can see in Figure~\ref{Fig:Single_Above} the flight path followed by the UAV during the mission. It takes off and flies to the first waypoint at $[-1.5,-3]$ before following a path to the second waypoint at $[-1.5,3]$. The red spherical target and blue pyramidal target are within both the camera field of view and the object detection algorithm's radius of interest while following this path. The UAV thus diverts to retrieve these targets. While the blue target is identified first, the forward motion of the vehicle results in the red target having closer proximity to the UAV as it enters \textit{Hover above target} mode. The red target is thus retrieved and deposited first.

The UAV then returns to the first waypoint and again follows the path to the second. This ensures that any additional targets along this path are detected. The blue target is again identified. It is retrieved and deposited at the drop site. The UAV again returns to the first waypoint and follows the path to the second waypoint. Finding no further targets, it proceeds to the third, fourth then fifth waypoints. While following the path to the sixth waypoint, it detects the green cuboidal target. It retrieves and deposits this target at the drop site. Having successfully deposited all three targets, the UAV returns to the landing site and the mission ends.

\begin{figure}
	\centering
	\includegraphics[scale=0.8]{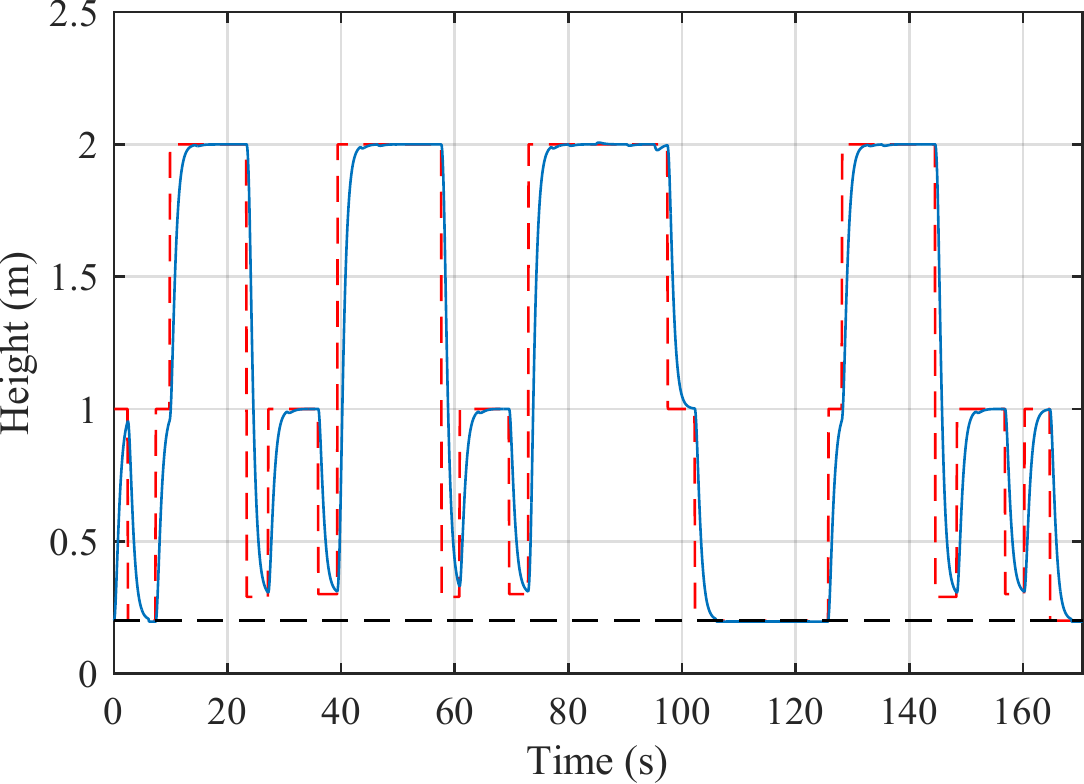}
	\caption{Height history during the mission. The dashed line indicates the height command.}
	\label{Fig:Single_Height}
\end{figure}

We can obtain further information on the mission by analysing the height history of the UAV, shown in Figure~\ref{Fig:Single_Height}. The UAV clearly begins the mission by taking off to a height of one metre. It then immediately lands again, indicating that a system fault has occurred, that is $(1 - P_s) > X \sim \mathcal{U}(0,1)$. The UAV then takes off again, before ascending to the search height of two metres and continuing the mission.

\begin{figure}
	\centering
	\includegraphics[scale=0.8]{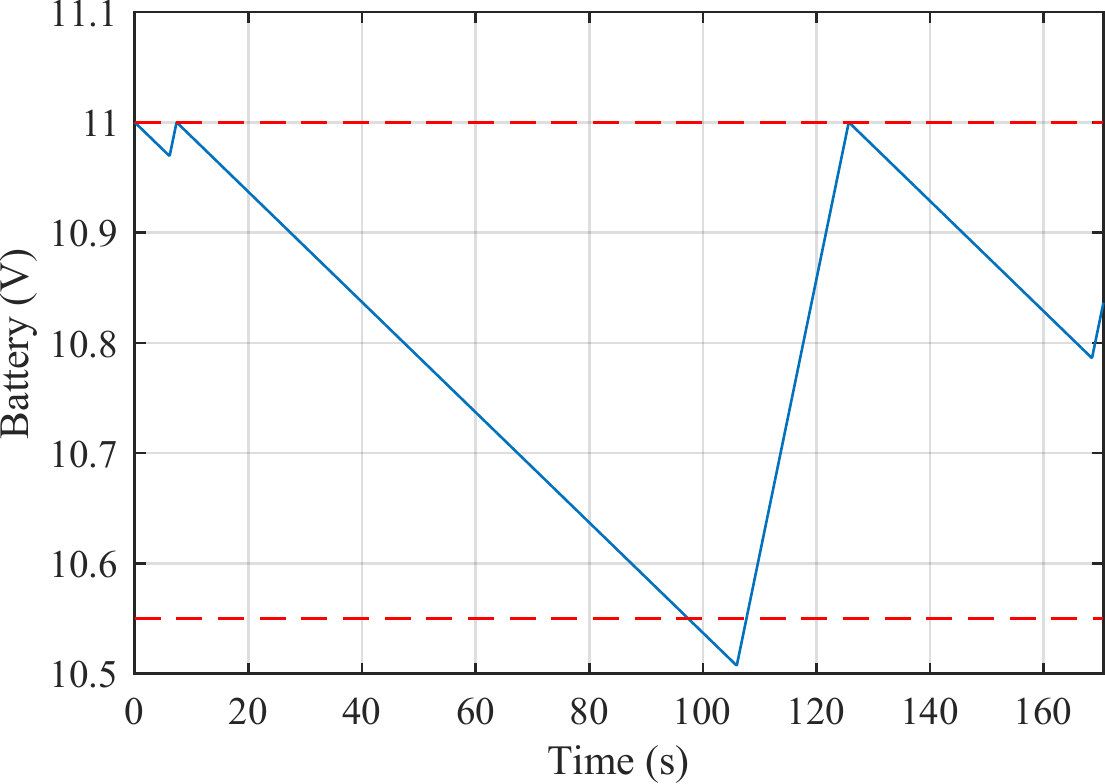}
	\caption{Battery voltage level during the mission. The upper and lower dashed lines indicate the maximum voltage level and warning threshold, respectively.}
	\label{Fig:Single_Battery}
\end{figure}

We can see that the UAV descends multiple times during the mission to an altitude of around 0.3 \si{\metre}. This is the height at which it grasps and drops the targets. It thus occurs six times, grasping and dropping each target. An additional descent occurs at around 100 \si{\second}. This is followed by a period of around 20 \si{\second} where the UAV rests on the floor. This is clearly an instance of a low battery warning triggering a return to base, where the UAV sits idle until it has recharged. It then resumes the mission. We may confirm this assumption by checking the battery voltage level history in Figure~\ref{Fig:Single_Battery}. The voltage level clearly crosses the threshold just prior to the 100 \si{\second} mark. The UAV then returns to base and recharges the battery until full, before resuming the mission. The mission is then completed before a second recharge is required.

\subsection{Monte Carlo Simulation}
We may approximate the probability of the mission being successful using Monte Carlo simulation. This provides a result which may be compared to the results of the model checking approach in future work. We may also analyse the paths taken by the finite state machine as the UAV performs the mission.

For the Monte Carlo simulation, we employ a simplified version of the simulation. The quadrotor initial position is fixed at $\vc{r}_{\qud,0} = [-1.8795, -2.5936, -0.2000]^T$ \si{\metre} with heading $ \psi_{\qud,0} = 0.9305 $ \si{\radian}. The drop site location is similarly fixed at $ \vc{r}_\ds = [1.3303, -1.3307, 0]^T $ \si{\metre}. The number of targets has been reduced to two. The initial positions of the targets are still randomly defined as previously described. The faults which may occur are still subject to the probabilities previously described. These simplifications are employed to facilitate comparison with the results of model checking the scenario in PRISM, which is currently in an earlier state of development.

The simulation is run for 2,000 iterations. The stochastic and probabilistic properties of the model are shown to impact a number of general mission properties. These properties include: the outcome of the mission; the duration of the mission; the times spent and distance travelled in each mode and; the transitions from one mode to another.

\subsubsection{Analyses of Mission Failures}
The probability of mission success is given in Table~\ref{Tab:Monte_Carlo_Results} as almost 80\%. We may examine missions for which the outcome was failure in order to obtain a number of reasons for these failures. First, at least one target may already be in the drop zone as the mission begins. Since the UAV ignores objects in the radius around the drop site, it is unable to find this target and reaches the end of the search path having relocated the remaining targets. Since \textit{all} targets must be retrieved for the mission to be considered successful, this instance results in a mission failure.

\begin{table}
	\centering
	\caption{Results of Monte Carlo method.}
	\label{Tab:Monte_Carlo_Results}
	\begin{tabular}{l r}
		\toprule
		Parameter & Occurrence \\
		\midrule
		Mission success & 0.8750 \\
		System fault in \textit{Initialise} mode & 0.0484 \\
		Actuator fault during mission & 0.0335 \\
		Grasping fault during target transportation & 0.0105 \\
		\bottomrule
	\end{tabular}
\end{table}

A similar outcome occurs when a target lands in the drop zone but does not register as having been deposited. This occurs when the target is dropped during \textit{Transport} and lands in the drop zone, rendering it invisible to the UAV. The premature release of the target can occur due to a grabber fault or a low battery warning.

Yet another route to mission failure is identified in the random positioning of the targets. One or more targets may be located such that they are ignored by the UAV, despite being outside of the drop zone. This may occur when targets are located in the corners of the environment. The target is outside of the object tracking algorithm's radius of interest $ R_c $ and, without any overlapping fly-bys occurring in this area, the target is ignored.

The remainder of mission failures occur due to actuator faults, which may occur at any time. Since returning to the landing site is a prerequisite for mission success, an emergency landing results in mission failure. This occurs regardless of whether all three targets have been retrieved and deposited at the drop site.

\subsubsection{Probabilities Derived from Simulation}
The probabilities of each of the three faults occurring are also given in Table~\ref{Tab:Monte_Carlo_Results}. A system fault may occur during \textit{Initialise} only. The probability of this occurring was given the explicit value $P_s = 0.05$. We use this probability as a verification check, with the result that the measured occurrence of a system fault in \textit{Initialise} is $ P_{s,m} = 0.0484 $. The result is found from over 4,000 samples, thus we can state with some confidence that a system fault occurs with expected frequency.

The actuator fault is specified as having a certain probability $P_a = 0.01 $ of occurring with a given time $T_a = 60$~\si{\second}. The results of the Monte Carlo experiment provide the occurrence of an actuator fault as percentage of the total number of iterations. As the actuator fault results in the mission ending in failure, it may only occur once per mission. An actuator fault is found to occur in 3.35\% of missions.

The grasping mechanism fault is similarly defined as having a probability  $P_g = 0.05$ of occurring in a $T_g = 60$~\si{\second} period. This fault can occur at any time during the mission. However, it only has impact when the mechanism is active, that is, when a target is tethered to the UAV. The occurrence of a grasping fault is thus limited to those modes during which a target is tethered to the UAV. A grasping mechanism fault is then found to cause a premature target drop in 1.05\% of the missions.

Mission time is found to have an average of 137.2~\si{\second}.

\begin{figure*}
	\centering
	\small
	\tikzstyle{line} = [draw, align=center, -latex', font=\scriptsize]
	\begin{tikzpicture}[node distance=2.2cm,on grid,auto,
		every state/.style={thick,font=\scriptsize}]


		\node[state,initial,accepting] (idle) [align=center] {1\\ Idle};
		\node[state] (takeoff) [above right of=idle, align=center] {2\\ Take-off};
		\node[state] (init) [below right of=takeoff, align=center] {3\\ Initialise};
		\node[state] (land) [below left of=init, align=center] {15\\ Land};
		\node[state] (search) [right of=init, align=center] {4\\ Search};
		\node[state] (identify) [above right of=search, align=center] {5\\ Identify};
		\node[state] (hover) [align=center, above right of=identify] {6\\ Hover\\ above};
		\node[state] (descendgrab) [align=center, right of=hover] {7\\ Descend\\ to grasp};
		\node[state] (grab) [below right of=descendgrab, align=center] {8\\ Grasp};
		\node[state] (ascend) [below right of=grab, align=center] {9\\ Ascend};
		\node[state] (transport) [below left of=ascend, align=center] {10\\ Transport};
		\node[state] (descenddrop) [align=center, below left of=transport] {11\\ Descend\\ to drop};
		\node[state] (drop) [left of=descenddrop, align=center] {12\\ Drop};
		\node[state] (rts) [align=center, below right of=search] {13\\ Return to\\ search};
		\node[state] (rtb) [align=center, below left of=rts] {14\\ Return\\ to base};
		\node[state] (reacquire) [align=center, right of=search, node distance=5cm] {16\\ Reacquire\\ target};
		\node[state] (emergency) [align=center, below of=idle, node distance=4cm] {17\\ Emergency\\ land};
		\node[state,dashed] (toemergency) [align=center, below of=emergency, node distance=3cm] {\{2,3,\ldots 16\}};
		\node[state,dashed] (tortb) [align=center, below of=rtb, node distance=3cm] {\{2,3,\ldots 13,16\}};

		\path[line]
		(idle) 			edge node {1} (takeoff)
		(takeoff) 		edge node {1} (init)
		(init) 			edge node {0.0484} (land)
						edge node {0.9516} (search)
		(search)		edge node [pos=0.85] {0.8505} (identify)
						edge [bend right] node [pos=0.5, left] {0.1460} (rtb)
		(identify)		edge node {1} (hover)
		(hover)			edge node {0.8685} (descendgrab)
						edge [bend right=45] node [pos=0.8, above left] {0.0751} (search)
		(descendgrab)	edge node {0.9732} (grab)
						edge [bend left=15] node [pos=0.4, above left] {0.0079} (search)
		(grab)			edge node {0.9998} (ascend)
		(ascend)		edge node {0.9739} (transport)
						edge node [above] {0.0020} (reacquire)
		(transport)		edge node {0.9601} (descenddrop)
						edge node {0.0033} (reacquire)
		(descenddrop)	edge node {0.9808} (drop)
						edge [bend right=15] node [below left, pos=0.3] {0.0016} (search)
		(drop)			edge node {0.5302} (rts)
						edge node {0.4698} (rtb)
		(rts)			edge node [pos=0.4] {0.9949} (search)
		(reacquire)		edge node [pos=0.2, above right] {0.8095} (hover)
						edge node [pos=0.5] {0.1905} (search)
		(rtb)			edge node [pos=0.5] {0.9997} (land)
		(land) 			edge node [pos=0.3] {0.9988} (idle)
		(emergency)		edge node {1} (idle)
		(toemergency)	edge node {} (emergency)
		(tortb)			edge node {} (rtb)
		;

	\end{tikzpicture}
	\caption{Finite-state machine showing probabilities of transitioning from one state to another.}
	\label{Fig:FSM_Probs}
\end{figure*}
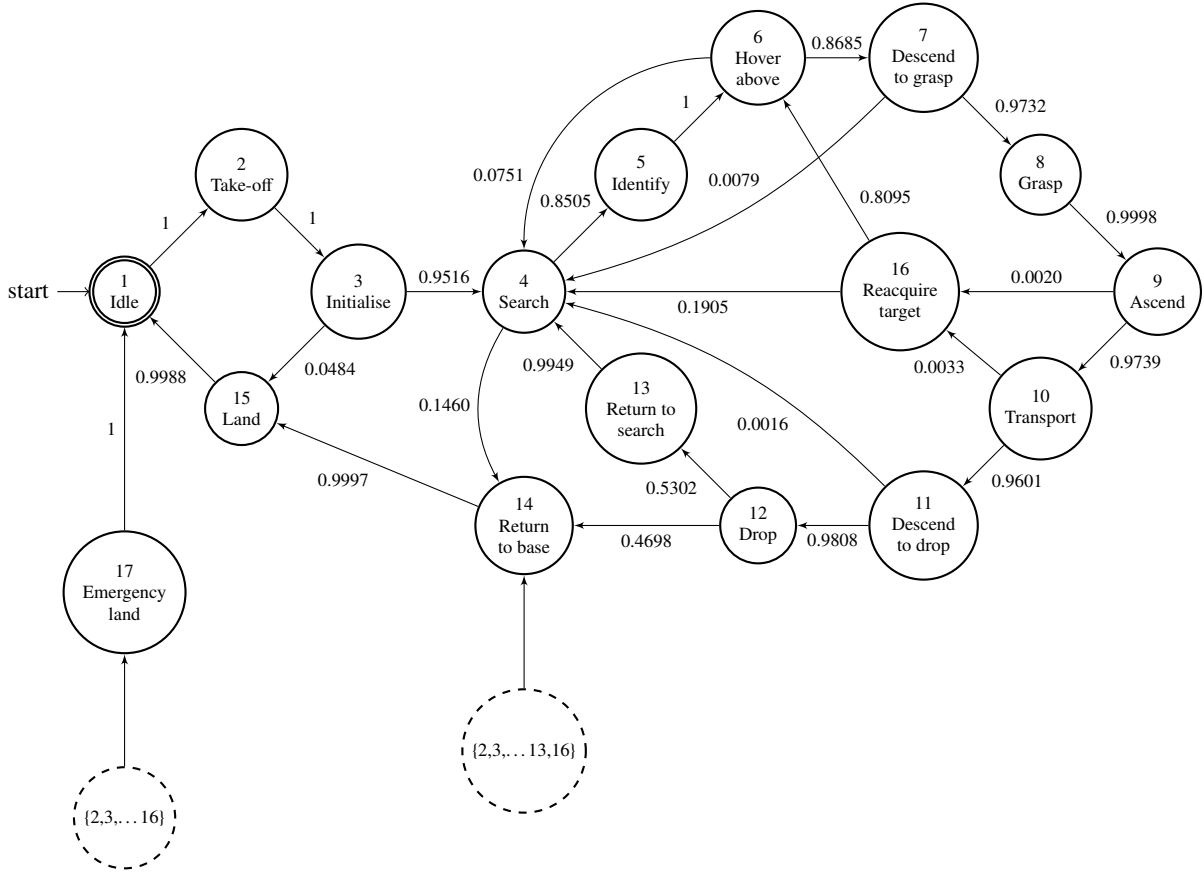

We may also consider the probabilities of progressing from one mode to another. The finite state machine in Figure~\ref{Fig:FSM} may be redrawn as Figure~\ref{Fig:FSM_Probs}, where the predicates for the transitions are replaced with the probabilities of a given transition occurring. Multiple modes may lead to both \textit{Emergency land} and \textit{Return to base}. The probabilities of these transitions occurring are listed separately in Table~\ref{Tab:Transition_Probs} for clarity.

\begin{table}
	\centering
	\caption{Probabilities of transitioning from a given state/mode to \textit{Emergency land} or \textit{Return to base}.}
	\label{Tab:Transition_Probs}
	\begin{tabular}{lrr}
		\toprule
		Mode 				& Emergency land & Return to base \\
		\midrule
		Idle 				& 0 		& 0 \\
		Take-off 			& 0 		& 0 \\
		Initialise 			& 0 		& 0 \\
		Search 				& 0.0036 	& 0.1460 \\
		Identify 			& 0 		& 0 \\
		Hover above target 	& 0.0025	& 0.0539 \\
		Descend to grasp 	& 0.0034 	& 0.0156 \\
		Grasp 				& 0 		& 0.0002 \\
		Ascend 				& 0.0017 	& 0.0224 \\
		Transport 			& 0.0010 	& 0.0356 \\
		Descend to drop 	& 0 		& 0.0176 \\
		Drop 				& 0 		& 0.4698 \\
		Return to search 	& 0 		& 0.0051 \\
		Return to base		& 0.0003 	& -- \\
		Land 				& 0.0012 	& -- \\
		Reacquire target 	& 0 		& -- \\
		Emergency land 		& -- 		& -- \\
		\bottomrule
	\end{tabular}
\end{table}

In the majority of cases, it is clear that there is a favoured mode to progress to from a given initial mode. Consider the transition from \textit{Initialise} to \textit{Search}, which has a much higher probability of occurring that a transition from \textit{Initialise} to \textit{Land}. In contrast, while \textit{Hover above target} is most likely to progress to \textit{Descend to grasp}, the probability of the UAV instead proceeding to \textit{Search} is non-negligible.

As expected, the probability of progressing from \textit{Drop} to \textit{Return to base} versus \textit{Return to search} is near equal. Since there are only two targets, the UAV will either \textit{Return to search} after dropping the first target or \textit{Return to base} after dropping the final target.

\section{Symbolic Abstract Model}
\label{Sec:Model_Checking}

As discussed in the introduction, the simulation model developed in this paper is intended to aid in the development of a framework for analysing autonomous systems and, in particular, the synthesis of high-level control for such systems. As an initial investigation, we are using the simulation model in the development and validation of a symbolic abstract model of the same scenario~\cite{Hoffmann2016}.  In the symbolic model the quantitative data for abstract modes and actions is derived from simulating small-scale models which, in turn, are derived from the presented simulation model. When building the abstraction non-determinism is introduced as the precise behaviour of the system is lost. The symbolic model is therefore represented as a Markov Decision Process (MDP) which exhibits both probabilistic and non-deterministic choice~\cite{Put94}.

An MDP consists of a set of states, set of actions, probabilistic transition function and reward function. In any state of an MDP there is a set of available actions (a subset of the action set) and for any state and available action the probabilistic transition function returns a distribution over successor states and the reward function returns the reward accumulated when taking the action.  The behaviour in any state consists of a non-deterministic choice from the set of available actions, followed by a probabilistic choice made according to the probabilistic transition function. An execution of an MDP is an sequence of such choices. The non-deterministic choices of available actions in each state are resolved by a strategy (also called an adversary or scheduler) which can base its choices on the execution up to that point. For a given strategy, the behaviour of an MDP is captured by a probability measure over the executions corresponding to the strategy's choices.

Key properties for MDPs are the probability of reaching a target set of states and the expected reward cumulated until this occurs. Of particular interest are the minimum and maximum values for these properties over all strategies. In models where the non-determinism in the MDP is through the abstraction of a concrete value, the minimum and maximum values yield upper and lower bounds on the precise value for the concrete system~\cite{FKNP11}.

\subsection{Model Checking with PRISM}

We have built and analysed the symbolic model using the probabilistic model checker PRISM~\cite{Kwiatkowska2011,Pri}.  Models in PRISM are expressed using a high level modelling language based on the Reactive Modules formalism \cite{AH99}. A model consists of a number of interacting modules. Each module consists of a number of finite-valued variables corresponding to the module's state and the transitions of a module are defined by a number of guarded commands of the form:
\[
\mathtt{[{<}action{>}]\ {<}guard{>}\ \rightarrow\ {<}prob{>}: {<}update{>} + \cdots + {<}prob{>}: {<}update{>}}
\]
A command consists of an (optional) action label, a guard and probabilistic choice between updates. A guard is a predicate over the variables of the modules, while an update specifies, using primed variables, how the variables of the module are updated when the command is taken. Interaction between modules is through both the guards (as guards can refer to variables of other modules) and action labels which allow modules to synchronise. Support for reward functions are through reward items of the form:
\[
\mathtt{[{<}action{>}]\ {<}guard{>} : {<}reward{>};}
\]
representing the reward accumulated when taken taking an action in a state satisfying guard. A reward is a real-valued expression which can include variables and constants of the model.

\begin{figure}[!t]
\centering
\input{search}
\caption{PRISM code relating to the abstract action search.}\label{fig:prismsearch}
\end{figure}

Our PRISM specification consists of four modules: a \textit{UAV} module, \textit{Time/Battery/Actuator} module, \textit{Movement} module and \textit{Object} module. The complete PRISM code can be found in the repository~\cite{repo}. To illustrate the abstract PRISM model, we present some fragments of the PRISM code corresponding to abstract actions representing search, grasp and release. In these code fragments the following variables appear:
\begin{itemize}
\item $\prismident{s1}$ representing the current mode of the UAV;
\item $\prismident{posx}$ and $\prismident{posy}$ representing the coordinates of UAV (the height of the UAV is abstracted);
\item $\prismident{retx}$ and $\prismident{rety}$ representing the position of the last way point;
\item $\prismident{c}$ representing the status of the actuator (operational or failed);
\item $\prismident{b}$ representing the battery charge level;
\item $\prismident{t}$ representing the current mission time;
\item $\prismident{NoOfObjs}$ representing the number of objects still to be found.
\end{itemize}
Fragments of the PRISM code for the abstract action search are presented in Figure~\ref{fig:prismsearch}. In the \textit{UAV} module, the UAV stays in the search mode if it has not found an object and has not reached the last possible coordinates (of the arena). If an object has been found, there is a transition to the identify mode. 

In the \textit{Time/Battery/Actuator} module, in accordance with the simulation model and small scale simulation results, the UAV moves one grid square per second, the rate of depletion of the battery is one unit per second and the probability of the actuator failing in one second is $0.00018$. If the actuator does not fail ($\prismident{c}{=}0$), the battery level and time are updated. In the abstract model when a transition corresponds to more than one second elapsing we have to ensure we have the correct values for the battery depletion and the cumulative probability of failure. More precisely, if a transition in the abstract model takes $t$ seconds, then when taking this transition the battery gets depleted by $t$ units and the probability that the actuator does not fail is $(1{-}\mathit{pf})^{t}$ (and hence the probability of failure is $1{-}(1{-}\mathit{pf})^{t}$).

\begin{figure}[!t]
\centering
\input{grab}
\caption{PRISM code fragments relating to the grasp and release abstract actions.}\label{fig:prismgrab}
\end{figure}

The fragments of the PRISM code for the three modules relevant to grab and release actions are presented in Figure~\ref{fig:prismgrab}.
The grasp and release abstract actions were both shown to take $0.01$ seconds via small-scale simulation. Through non-determinism (see Figure~\ref{fig:prismgrab}), the PRISM model abstracts these timing values to lower and upper bounds ($0$ and $1$). 
In the \textit{Object} module the number of non-deposited objects $\prismident{NoOfObjs}$ is decremented when an object is released at the depot site. 

\begin{figure}
\centering
\input{properties}
\caption{PRISM properties code.}\label{fig:prismprops}
\end{figure}

We evaluated the symbolic implementation by building a model for each combination of object positioning in the arena, while keeping the base and the depot sites fixed. The properties that we focused on determine probability bounds for a mission to be successful, and for a mission to end due to an actuator fault respectively. The PRISM properties are shown in  Figure~\ref{fig:prismprops} and we have computed minimum and maximum values yielding upper and lower bounds for the concrete system. The bounds obtained are:
\begin{itemize}
\item $[0.661,0.891]$ for the probability of successful mission completion (the simulation results returned $0.8750$);
\item $[0.018,0.041]$ for the probability of an actuator fault (the simulation results returned $0.0335$);
\item $[101.3,223.7]$ for the expected mission time (the simulation results returned $137.2$).
\end{itemize}
As can be seen the bounds from the symbolic model encompass the outcomes from the Monte Carlo simulations presented in Section~\ref{Sec:Results}.
We can also see that the relative differences of the bounds are of the same order of magnitude. Although the bounds are not tight, they are promising for an initial investigation. One potential reason for the bounds not being tight is how time has been abstracted in the model. In particular, in some cases the lower bound on the time taken for some transitions is 0 (as can be seen in Figure~\ref{fig:prismgrab}), allowing certain abstract actions to be taken repeatedly without influencing the overall mission time. This behaviour is clearly not realisable in the simulation model.

\section{Conclusions}
\label{Sec:Conclusions}
We may draw some conclusions on the verification of real-world autonomous systems from both the development of our continuous-time model and the results of the simulation.

\subsection{Limitations of the Model}
As an entity in simulation only, our model simplifies or omits some complex real-world behaviours. First, target impact on UAV behaviour is not modelled. In reality, grasping the target would alter the mass and moments of inertia of the UAV. This would then affect the closed-loop response of the vehicle, potentially changing the outcome of the mission. Specifically, the impact of the target mass on the state feedback controller must be considered in any practical implementation of the scenario.

Additionally, no fault detection algorithm is implemented. Faults in the system are assumed to be detected instantaneously. The autonomous controller then switches to an appropriate mode to mitigate the effects of the fault. In reality, the faults would be detected either be sensors monitoring the affected components or by any simple fault detection method \cite{Isermann2006}.

The lack of disturbances and the use of feedback linearisation in the controller precludes the need for integral action. This is because the linearising feedbacks perfectly cancel out the non-linearities and system properties in simulation \cite{Ireland2015a}. In reality, this feedbacks do not perfectly cancel out the system dynamics. Thus, there is a need for a more robust controller. The presence of disturbances in reality further establishes the need for integral action.

While a simple power model is implemented in our quadrotor agent, its role is simply as an additional predicate for a greater variety of transitions in the finite state machine. In reality, a decreasing battery voltage may impact the performance of the rotors and can affect the flight characteristics of the UAV even before a return to base command is issued.

The camera sensor and grasping mechanism are optimally positioned such that an object centred in the camera image is directly below the grasping mechanism. In reality, the grasper would obscure the object from view. In implementing the autonomous quadrotor system, we would need to carefully consider the positions of the camera and grasper and their impact on the target detection and grasping.

Finally, we have simplified the interaction and operation of the UAV's subsystems by assuming they function flawlessly. For example, the camera sensor is assumed to have a high framerate and perceives the targets as solid blocks of colour. The object detection module is assumed to execute at a high bandwidth, resulting in no lag between the camera and the eventual centroid coordinates. In reality, the object detection will likely be imperfect and will run at a relatively low bandwidth to ensure real-time execution. This would then impact both the visual controller and the autonomous mode switching.

\subsection{Limitations of the Autonomous Controller}
The modes of the autonomous controller FSM were primarily determined prior to implementation. Early simulation testing resulted in the occurrence of events for which the FSM had not been programmed to consider. Some additional modes were thus incorporated into the FSM, such as \textit{Reacquire target} or \textit{Return to search}. Experimental testing in the future is likely to highlight further gaps in the FSM and require the addition of more guidance modes.

The current implementation of the autonomous controller is deliberately lacking in parts, to better facilitate the chance of mission failures. Analysis of the mission specifics for each of the 2,000 iterations provides a number of reasons for mission failure.

First, there is the possibility that the UAV believes the mission has failed when it has in fact been successful. This may occur when the location of one of the targets is randomly defined to be already in the specified radius around the drop site. The UAV is unable to detect this target. It therefore follows the mission in depositing the remaining targets, before reaching the end of the search path and returning to base, believing the mission has failed. This outcome is a consequence of the fact that both target location and drop site are defined randomly and may therefore coincide. The autonomous behaviour of the UAV does not conceive of this occurrence, knowing only that there are three targets and that it has deposited two.

A similar outcome can occur when all targets are initially outside of the drop site radius, but at least one is somehow undetectable to the UAV. This only occurs when the target is positioned in the corner of the environment, as the overlap in sensor perception elsewhere prevents this occurrence otherwise.

\subsection{Future Work}
We aim to continue this work on two fronts. First, we are continuing our work in formal verification of autonomous systems in PRISM. The symbolic model of our system must satisfy the dual requirements of accurately describing the intricacies of the system, while also being sufficiently simple that it may be verified in a short period of time. The pursuit of this research is intended to lead to the creation of a framework for the verification of autonomous systems.
For that we need to formally prove a connection between the simulation and symbolic models which will allow us to infer results for the actual system from those obtained through model checking the symbolic model. The longer term goal is to develop a highly adaptable framework for analysing autonomous systems. For example, once the initial small-scale simulation models for abstract actions and symbolic models have been set up for a certain scenario, different decision algorithms can be easily substituted and analysed. In addition, we can reuse the quantitative data for abstract actions obtained from the small scale simulation models when analysing different scenarios in which the same abstract actions occur.

On the second front, we are implementing our autonomous mission in controlled laboratory conditions. By doing this, we hope to identify additional factors which affect the mission outcome and determine whether these factors may be mitigated by robust system design or by real-time system verification. We are also investigating the required hardware to support real-time verification and onboard autonomy.


\section*{Acknowledgments}
This work was supported by the Engineering and Physical Sciences Research Council [grant number EP/N508792/1].

\bibliographystyle{unsrt}
\bibliography{Bibliography}

\begin{thebibliography}{10}

\bibitem{Webster2011}
M.~Webster, M.~Fisher, N.~Cameron, and M.~Jump.
\newblock {Formal Methods for the Certification of Autonomous Unmanned Aircraft
  Systems}.
\newblock {\em Computer Safety, Reliability, and Security}, 6894:228--242,
  2011.

\bibitem{Lerda2008}
F.~Lerda, J.~Kapinski, H.~Maka, E.~Clarke, and B.~Krogh.
\newblock {Model Checking In-The-Loop}.
\newblock In {\em Proceedings of the American Control Conference}, pages
  2734--2740, 2008.

\bibitem{Kwiatkowska2011}
M.~Kwiatkowska, G.~Norman, and D.~Parker.
\newblock {PRISM 4.0: Verification of probabilistic real-time systems}.
\newblock In G~Gopalakrishna and S~Qadeer, editors, {\em Proceedings of the
  23rd International Conference on Computer Aided Verification (CAV'11)},
  volume 6806 of {\em LNCS}, pages 585--591. Springer, Jul 2011.

\bibitem{Abbas2013}
H~Abbas, G~Fainekos, S~Sankaranarayanan, F~Ivancic, and A~Gupta.
\newblock {Probabilistic Temporal Logic Falsification of Cyber-Physical
  Systems}.
\newblock {\em ACM Transactions on Embedded Computing Systems}, 12(2):30, 2013.

\bibitem{Sankaranarayanan2012}
S.~Sankaranarayanan and G.~Fainekos.
\newblock {Falsification of Temporal Properties of Hybrid Systems Using the
  Cross-Entropy Method}.
\newblock In {\em Proceedings of the 15th International Conference on Hybrid
  Systems: Computation and Control}, pages 125--134, Beijing, apr 2012.

\bibitem{Nghiem2010}
T.~Nghiem, S.~Sankaranarayanan, G.~Fainekos, F.~Ivancic, A.~Gupta, and
  G.~Pappas.
\newblock {Monte-carlo techniques for falsification of temporal properties of
  non-linear hybrid systems}.
\newblock In {\em Proceedings of the 13th International Conference on Hybrid
  Systems: Computation and Control}, pages 211--220, Stockholm, apr 2010.

\bibitem{Althoff2014}
M.~Althoff and .~Dolan.
\newblock {Online verification of automated road vehicles using reachability
  analysis}.
\newblock {\em IEEE Transactions on Robotics}, 30(4):903--918, 2014.

\bibitem{Gillula2010}
J.~Gillula, H.~Huang, M.~Vitus, and C.~Tomlin.
\newblock {Design of guaranteed safe maneuvers using reachable sets: Autonomous
  quadrotor aerobatics in theory and practice}.
\newblock In {\em Proceedings - IEEE International Conference on Robotics and
  Automation}, pages 1649--1654, 2010.

\bibitem{Althoff2015}
D.~Althoff, M.~Althoff, and S.~Scherer.
\newblock {Online Safety Verification of Trajectories for Unmanned Flight with
  Offline Computed Robust Invariant Sets}.
\newblock In {\em Proceedings of the IEEE/RSJ International Conference on
  Intelligent Robots and Systems}, pages 3470--3477, Hamburg, sep 2015. IEEE.

\bibitem{Bochot2009}
T.~Bochot, P.~Virelizier, H.~Waeselynck, and V.~Wiels.
\newblock {Model Checking Flight Control Systems: The Airbus Experience}.
\newblock In {\em 31st International Conference on Software Engineering - ICSE
  '09}, pages 18--27, Vancouver, May 2009. IEEE.

\bibitem{Henzinger1996}
T.~Henzinger.
\newblock {The Theory of Hybrid Automata}.
\newblock In {\em Proceedings 11th Annual IEEE Symposium on Logic in Computer
  Science}, pages 278--292, New Brunswick, NJ, Jul 1996. IEEE.

\bibitem{Chutinan2000}
A.~Chutinan and B.~Krogh.
\newblock {Verification of Infinite-State Dynamic Systems Using Approximate
  Quotient Transition Systems}.
\newblock {\em IEEE Transactions on Automatic Control}, 46(9):101--109, Sep
  2000.

\bibitem{Kowalewski1999}
S~Kowalewski, S~Engell, J~Preu{\ss}ig, and O~Stursberg.
\newblock {Verification of logic controllers for continuous plants using timed
  condition/event-system models}.
\newblock {\em Automatica}, 35(3):505--518, 1999.

\bibitem{Bohn2007}
C.~Bohn.
\newblock {Heuristics for Designing the Control of a UAV Fleet with Model
  Checking}.
\newblock {\em Lecture Notes in Economics and Mathematical Systems},
  588:21--36, 2007.

\bibitem{Konur2012}
S.~Konur, C.~Dixon, and M.~Fisher.
\newblock {Analysing robot swarm behaviour via probabilistic model checking}.
\newblock {\em Robotics and Autonomous Systems}, 60(2):199--213, 2012.

\bibitem{Liu2007}
W.~Liu, A.T Winfield, and J.~Sa.
\newblock {Modelling Swarm Robotic Systems : A Case Study in Collective
  Foraging}.
\newblock In {\em Proceedings of Towards Autonomous Robotic Systems}, pages
  25--32, Aberystwyth, Sep 2007.

\bibitem{Mikael2012}
L.~Mika{\"{e}}l.
\newblock {\em {Formal Verification of Flexibility in Swarm Robotics}}.
\newblock Masters thesis, Universit{\'{e}} libre de Bruxelles, 2012.

\bibitem{Humphrey2013}
L.~Humphrey.
\newblock {Model Checking for Verification in UAV Cooperative Control
  Applications}.
\newblock In Fariba Fahroo, Le~Yi Wang, and George Yin, editors, {\em Recent
  Advances in Research on Unmanned Aerial Vehicles}, chapter~4. Springer, 2013.

\bibitem{Choi2012}
J.~Choi.
\newblock {\em {Model Checking for Decision Making Behaviour of Heterogeneous
  Multi-Agent Autonomous System}}.
\newblock Phd thesis, Cranfield University, 2012.

\bibitem{Kubera2010}
Y.~Kubera, P.~Mathieu, and S.~Picault.
\newblock {Everything can be Agent! (Extended Abstract)}.
\newblock In {\em Proceedings of the 9th International Conference on Autonomous
  Agents and Multiagent Systems (AAMAS'10)}, pages 1547--1548, Toronto, May
  2010.

\bibitem{Frank2001}
A.~Frank, S.~Bittner, and M.~Raubal.
\newblock {Spatial and Cognitive Simulation with Multi-agent Systems}.
\newblock In Daniel~R Montello, editor, {\em Spatial Information Theory:
  Foundations of Geographic Information Science}, pages 124--139. Springer
  Berlin Heidelberg, 2001.

\bibitem{Bouabdallah2004}
S.~Bouabdallah, P.~Murrieri, and R.~Siegwart.
\newblock {Design and Control of an Indoor Micro Quadrotor}.
\newblock In {\em Proceedings of IEEE International Conference on Robotics and
  Automation}, pages 4393--4398. IEEE, 2004.

\bibitem{Voos2009}
H.~Voos.
\newblock {Nonlinear Control of a Quadrotor Micro-UAV Using
  Feedback-Linearization}.
\newblock In {\em Proceedings of the 2009 IEEE International Conference on
  Mechatronics}, Malaga, 2009. IEEE.

\bibitem{Ireland2014}
M.~Ireland.
\newblock {\em {Investigations in Multi-Resolution Modelling of the Quadrotor
  Micro Air Vehicle}}.
\newblock {PhD}, University of Glasgow, may 2014.

\bibitem{Tayebi2006}
A.~Tayebi and S.~McGilvray.
\newblock {Attitude Stabilization of a VTOL Quadrotor Aircraft}.
\newblock {\em IEEE Transactions on Control Systems Technology},
  14(3):562--571, may 2006.

\bibitem{Mueller2014}
M.~Mueller and R.~D'Andrea.
\newblock {Stability and control of a quadrocopter despite the complete loss of
  one, two, or three propellers}.
\newblock In {\em Proceedings of the 2014 IEEE International Conference on
  Robotics and Automation (ICRA)}, pages 45--52, Hong Kong, may 2014. Ieee.

\bibitem{Hartley2003}
R~Hartley and A~Zisserman.
\newblock {\em {Multiple View Geometry in Computer Vision}}.
\newblock Cambridge University Press, Cambridge, 2nd edition, 2003.

\bibitem{Ireland2015}
M.~Ireland and D.~Anderson.
\newblock {An Investigation of the Effects of Model Resolution on Control of a
  Quadrotor Micro Air Vehicle}.
\newblock {\em IJUSEng}, 3(1):17--25, 2015.

\bibitem{Ireland2015a}
M.~Ireland, A.~Vargas, and D.~Anderson.
\newblock {A Comparison of Closed-Loop Performance of Multirotor Configurations
  Using Non-Linear Dynamic Inversion Control}.
\newblock {\em Aerospace}, 2(2):325--352, 2015.

\bibitem{Das2009}
A.~Das, K.~Subbarao, and F.~Lewis.
\newblock {Dynamic inversion with zero-dynamics stabilisation for quadrotor
  control}.
\newblock {\em Control Theory {\&} Applications, IET}, 3(3):303--314, 2009.

\bibitem{Hoffmann2016}
R.~Hoffmann, M.~Ireland, A.~Miller, G.~Norman, and S.~Veres.
\newblock {Autonomous Agent Behaviour Modelled in PRISM: A Case Study}.
\newblock In {Bonaki, D and Wijs, A}, editor, {\em {23rd International SPIN
  Symposium on Model Checking of Software}}, volume 9641 of {\em LNCS},
  Eindhoven, April 2016.
\newblock [Accepted for publication].

\bibitem{Put94}
M.~Puterman.
\newblock {\em Markov Decision Processes: Discrete Stochastic Dynamic
  Programming}.
\newblock John Wiley and Sons, 1994.

\bibitem{FKNP11}
V.~Forejt, M.~Kwiatkowska, G.~Norman, and D.~Parker.
\newblock Automated verification techniques for probabilistic systems.
\newblock In M.~Bernardo and V.~Issarny, editors, {\em Formal Methods for
  Eternal Networked Software Systems (SFM'11)}, volume 6659 of {\em LNCS},
  pages 53--113. Springer, 2011.

\bibitem{Pri}
{PRISM} web site.
\newblock www.prismmodelchecker.org.

\bibitem{AH99}
R.~Alur and T.~Henzinger.
\newblock Reactive modules.
\newblock {\em Formal Methods in System Design}, 15(1):7--48, 1999.

\bibitem{repo}
Abstract {PRISM} model.
\newblock BitBucket Repository, 2016.
\newblock bitbucket.org/ruthhoffmann/symbolic-uav-model.

\bibitem{Isermann2006}
R.~Isermann.
\newblock {\em {Fault-Diagnosis Systems}}.
\newblock Springer, 2006.

\end{thebibliography}

\end{document}